\documentclass[10pt,twocolumn,letterpaper]{article}

\usepackage[pagenumbers]{cvpr} 

\usepackage[table]{xcolor}
\usepackage[rightcaption]{sidecap}
\usepackage{multirow}
\usepackage{makecell}
\usepackage{algorithm}
\usepackage{algpseudocode}

\DeclareMathOperator*{\argmax}{arg\,max}

\newcommand{\TODO}[1]{\textbf{\color{red}[TODO: #1]}}
\renewcommand{\TODO}[1]{}

\definecolor{cvprblue}{rgb}{0.21,0.49,0.74}
\usepackage[pagebackref,breaklinks,colorlinks,allcolors=cvprblue]{hyperref}

\title{Subtract the Corruption: Training-Data-Free Corrective Machine Unlearning using Task Arithmetic}

\author{
Mostafa Mozafari \quad
Farooq Ahmad Wani \quad
Maria Sofia Bucarelli \quad
Fabrizio Silvestri\\[2pt]
Sapienza University of Rome\\[2pt]
{\tt\small mostafamozafari1996@gmail.com, farooqahmad.wani@uniroma1.it}\\
{\tt\small mariasofia.bucarelli@uniroma1.it, fsilvestri@diag.uniroma1.it}
}

\begin{document}
\maketitle
\begin{abstract}
Corrupted training data are ubiquitous. Corrective Machine Unlearning (CMU) seeks to remove the influence of such corruption post-training. Prior CMU typically assumes access to identified corrupted training samples (a ``forget set''). However, in many real-world scenarios the training data are no longer accessible. We formalize \emph{source-free} CMU, where the original training data are unavailable and, consequently, no forget set of identified corrupted training samples can be specified. Instead, we assume a small proxy (surrogate) set of corrupted samples that reflect the suspected corruption type without needing to be the original training samples. In this stricter setting, methods relying on forget set are ineffective or narrow in scope. We introduce \textit{Corrective Unlearning in Task Space} (CUTS), a lightweight weight space correction method guided by the proxy set using task arithmetic principles. CUTS treats the clean and the corruption signal as distinct tasks. Specifically, we briefly fine-tune the corrupted model on the proxy to amplify the corruption mechanism in the weight space, compute the difference between the corrupted and fine-tuned weights as a proxy task vector, and subtract a calibrated multiple of this vector to cancel the corruption. Without access to clean data or a forget set, CUTS recovers a large fraction of the lost utility under label noise and, for backdoor triggers, nearly eliminates the attack with minimal damage to utility, outperforming state-of-the-art specialized CMU methods in source-free setting. Code is available at \url{https://github.com/mosix11/CUTS}.

\end{abstract}
    
\section{Introduction}
\label{sec:intro}

\begin{figure}
    \centering
    \includegraphics[width=\linewidth]{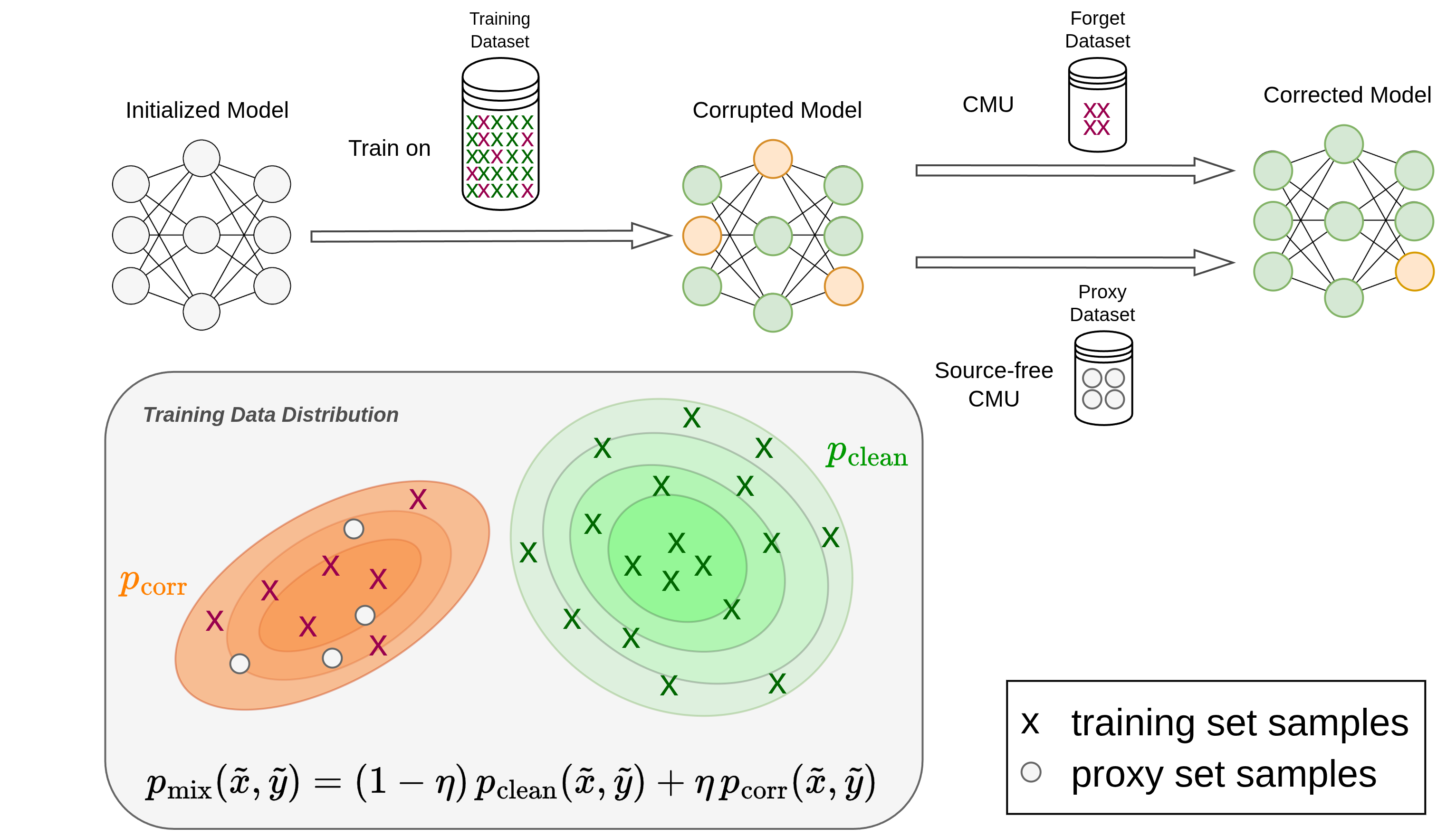}
    \caption{The training distribution mixes clean and corrupted distributions. Classic CMU corrects via an identified \textit{forget set} of corrupted training samples. In source-free setting, the training dataset is inaccessible; one can only use a \textit{proxy set} of samples from the corruption distribution not necessarily seen during training.}
    \label{fig:setting_intro}
\end{figure}

Deep neural networks owe much of their success to training on massive and diverse datasets with human or automated annotations \cite{DualT}. However, such datasets inevitably contain corrupted samples due to labeling mistakes, web-scale collection artifacts, or adversarial manipulations \cite{6685834, 9423393, cheng2021learning, NPC, canmachinelearningbesecure, 10646610, Liu2018TrojaningAO, Chen2017TargetedBA, song2022learningnoisylabelsdeep}. As dataset sizes grow, ensuring perfect data integrity becomes unrealistic, and modern overparameterized models tend to memorize or amplify the effects of these corruptions \cite{arpit2017closerlookmemorizationdeep, northcutt2021pervasive, goel2024correctivemachineunlearning, 10.5555/3294996.3295110}. A large body of work attempts to prevent corruption during data curation or training. For label noise, methods include \textit{noise cleansing} \cite{10.5555/3294771.3294863, 8578680, coteaching, 9156369, FINEsamples} and \textit{noise-robust learning} with robust losses \cite{GCE, SCE, ma2020normalized, farooqpaper} or regularization \cite{earlylearning, robustearlylearning}. Similarly, for backdoor attacks, analogous up-front defenses include pre-training data sanitization as well as training procedures that are robust to triggered training samples \cite{spectralsignatures, antibackdoorlearning, huang2022backdoor, 10.5555/3294996.3295110, nguyen2021wanet, 9060997, backdoorllmsurvey}. Yet these approaches rely on access to the full training set and are inapplicable once the model has already been trained. In many real-world scenarios, the corruption is discovered only after training or deployment and retraining might be costly and time consuming. This motivates post-training \emph{Corrective Machine Unlearning} (CMU) \cite{goel2024correctivemachineunlearning}, which aims to remove the detrimental effects of corrupted data from a trained model with minimal data, computation, and downtime. Unlike \emph{Privacy-Preserving Unlearning} (\cite{ginart2019makingaiforgetyou, certifieddataremoval, bourtoule2020machineunlearning, Xu2023MachineUS}), CMU does not require per-sample deletion guarantees, and targets the broader harmful effects of corruption.
\looseness=-1

In real-world deployments the original training data are often inaccessible or too costly to reproduce due to privacy, licensing, or operational constraints. However, post-deployment monitoring, red-teaming, or incident reports can reveal patterns consistent with specific classes of corruption \cite{goel2024correctivemachineunlearning}. To address this, we introduce and formalize the \emph{source-free} CMU setting, where only the trained model and an identified corruption type are available, without access to any original training data (\cref{fig:setting_intro}). Then, a practitioner compiles a small \emph{proxy} (surrogate) dataset drawn from the same input domain or a close approximation (e.g., collected unlabeled samples, inference data, or generated data) and instantiates the suspected corruption on this proxy, such as injecting label noise or trigger patterns. 
Critically, this setup imposes  weaker requirements than the classic CMU paradigm, which requires access to the actual corrupted training samples, necessitating storage of original data and identification of (a subset of) exact corrupted instances. 
\looseness=-1

\begin{figure}[h!]
    \centering
    \includegraphics[width=\linewidth]{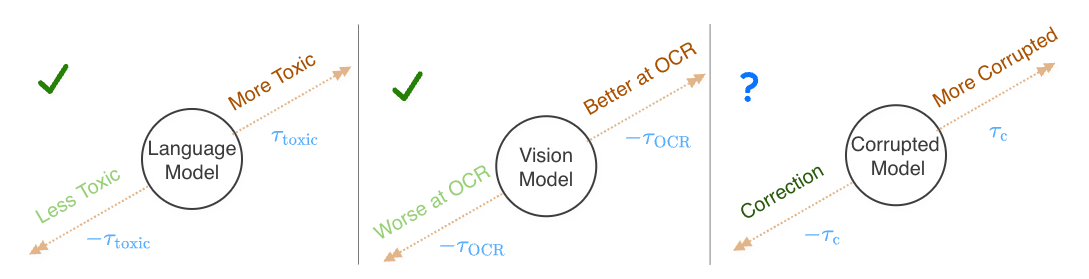}
    \caption{Task vector subtraction from a pre-trained model induces partial task forgetting, as demonstrated for partial unlearning of toxic language and degrading OCR performance in \cite{ilharco2023editing}. Our motivation is the following: if corruption can be modeled as a task, task arithmetic enables source-free, proxy-guided CMU.}
    \label{fig:intro_question}
\end{figure}

\emph{Task Arithmetic} \cite{ilharco2023editing} has been shown effective in model editing. Adding task vectors to a pre-trained model can approximate multi-task behavior, and negating them can induce partial forgetting of the tasks. Our key hypothesis is that training on partially corrupted data can be viewed as a two-task learning problem; a clean task and a corruption task. The total weight update is thus approximately the sum of two task vectors in weight space. If we can estimate the corruption task vector using a proxy dataset, we can subtract it from the trained weights to cancel the corruption effect (\cref{fig:intro_question}). This builds on the observed local linearity and additivity of task vectors in pre-trained models \cite{ortiz-jimenez2023task, zeng2025efficientmodeleditingtask, li2025when, iurada2025efficient}.
Practically, we fine-tune the corrupted model on the proxy with simulated corruption, extract the weight difference as the corruption task vector, then apply a scaled negative edit. This lightweight, source-free procedure effectively mitigates the detrimental impact of symmetric/asymmetric label noise \cite{asym-noise-def} and BadNets-style triggers \cite{badnets}.
\looseness=-1



\paragraph{Contributions.}
\begin{itemize}
    \item We formalize \emph{source-free Corrective Machine Unlearning}, where the goal is to correct corrupted models using only a small \emph{proxy} dataset of corrupted samples, without access to the original training data and forget set.      
    
    \item We propose a post-training corrective method (CUTS) based on \textit{task arithmetic}, treating corruption itself as a task. By estimating a corruption task vector using the proxy dataset, we show that the effect of corruption in corrupted models is encoded along an approximately linear direction in weight space, largely but not entirely distinct from the clean signal.

    \item We demonstrate that CUTS is effective across a spectrum of corruption regularity, from unstructured corruptions (e.g., random label noise, which induces per-sample memorization) to structured ones (e.g., backdoor triggers with consistent learnable patterns), by targeting and counteracting corruption approximated directions in weight space rather than per-sample associations.
    

\end{itemize}

We conduct extensive experiments across architectures (pre-trained and randomly initialized), datasets, and corruption types, providing a systematic analysis of our method’s effectiveness, generalization, and limitations.

\section{Related Works}
\label{sec:rel_works}

\paragraph{Regularity and the Interplay Between Corruption and Memorization.}
Deep neural networks are known to memorize arbitrary data \citep{arpit2017closerlookmemorizationdeep, 10.1145/3446776}. This memorization supports learning from rare or atypical examples \citep{feldman2021doeslearningrequirememorization, feldman2020neuralnetworksmemorizewhy}. Contrary to earlier assumptions \citep{50814, stephenson2021geometrygeneralizationmemorizationdeep}, memorization is not confined to specific layers but distributed across neurons throughout the network,  and memorizing mislabeled and hard samples shows similar dynamics \citep{maini2023neuralnetworkmemorizationlocalized}.
\looseness=-1

Regularity refers to the degree of self-similarity among corrupted examples \cite{REM}.
Low-regularity corruptions (e.g., random label noise) lack structure and force the network to rely on per-sample memorization, while high-regularity corruptions (e.g., spatial backdoor triggers) provide consistent patterns that are easily learned and generalized. 
Consequently, memorization burden increases with decreasing regularity \citep{REM, maini2023neuralnetworkmemorizationlocalized}. 
This distinction underlines that memorization and corruption are intertwined, but their relationship depends critically on the regularity of the corruption.
\looseness=-1

\paragraph{Corrective Machine Unlearning.}
Machine unlearning seeks to remove the influence of specific training data from a model, ideally as if those samples had never been seen \citep{7163042}. 
Early work motivated by privacy regulations proposed exact or approximate forgetting mechanisms \citep{ginart2019makingaiforgetyou, bourtoule2020machineunlearning, certifieddataremoval}, later expanded through influence-based updates \citep{golatkar2020eternalsunshinespotlessnet} and distillation methods. 
Recent surveys \citep{wang2024machineunlearningcomprehensivesurvey, Xu2023MachineUS} classify approaches by efficiency and privacy fidelity. 
Beyond privacy, unlearning has become essential for removing corrupted or mislabeled data discovered post-training. 
Goel et al. \cite{goel2024correctivemachineunlearning} formalized \textit{Corrective Machine Unlearning} (CMU), which aims to mitigate the impact of corrupted samples while preserving model utility. 
Unlike classical unlearning that assumes all deletion data are known, CMU methods must operate under partial discovery and diverse corruption types. 
Studies show that existing methods perform inconsistently across corruption regularities, motivating methods that adaptively handle a wider range of corruptions \citep{goel2024correctivemachineunlearning, REM}. 
However, the majority of prior CMU approaches \cite{goel2024correctivemachineunlearning, SSD, SCRUB, BadT, REM} require access to some identified corrupted samples from the original training data or the entire training data, making them impractical when data are unavailable due to legal or operational constraints. Our source-free formulation relaxes these assumptions. Among existing CMU methods, SAP \cite{kodge2025sap} (for label noise) and Potion \cite{potion} (for poison triggers) are specialized SOTA approaches that can be adapted to source-free setting, and \cite{CF} provides a trivial catastrophic forgetting baseline via fine-tuning on a small clean set.
\looseness=-1

\paragraph{Task Arithmetic for Model Editing.}
Task arithmetic \cite{ilharco2023editing} offers a lightweight, data-free paradigm for model editing by operating on task vectors in the weight space. Model behaviors can be composed, removed, or transferred through linear combinations of task-specific vectors, enabling fast adaptation without joint fine-tuning.
The success of task arithmetic has been linked to \textit{weight disentanglement}; the property that different task directions act on largely disjoint functional regions, minimizing interference between edits \citep{ortiz-jimenez2023task}. 
Recent works have also begun to establish theoretical foundations for task arithmetic, explaining when and why it succeeds in nonlinear settings \citep{li2025when, zeng2025efficientmodeleditingtask, iurada2025efficient}, although these analyses often rely on assumptions, such as smooth loss landscapes or small weight updates, that may not strictly hold in practical scenarios.
\looseness=-1
\section{Preliminaries}
\label{sec:preliminaries}
\subsubsection*{Problem Definition}
We consider image classification where the input space $\mathcal{X}$ is continuous (images) and the label space is discrete $\mathcal{Y}=\{1,\dots,K\}$. Let $p_{\text{clean}}$ denote the clean data distribution. A (possibly stochastic) corruption kernel $\kappa(x',y'\mid x,y)$ specifies how a clean pair $(x,y)$ is transformed into a corrupted pair $(x',y')$ if corruption occurs. Pushing $p_{\text{clean}}$ through $\kappa$ yields the corrupted distribution 
\begin{equation}
    p_{\text{corr}}(x',y') = \sum_{y=1}^K \int_{\mathcal{X}} p_{\text{clean}}(x,y)\;\kappa(x',y'\mid x,y)\;dx.
    \label{eq:corrupted_marginal}
\end{equation}
The observed training distribution by the learning algorithm mixes clean and corrupted distributions, it is formulated as \begin{equation}
    p_{\text{mix}}(\tilde x,\tilde y)
    =
    (1-\eta)\,p_{\text{clean}}(\tilde x,\tilde y)
    +
    \eta\,p_{\text{corr}}(\tilde x,\tilde y).
    \label{eq:mixture_distribution}
\end{equation}

The training dataset $\mathcal{D}_{\text{train}}{=}\{(\tilde{x}_i,\tilde{y}_i)\}_{i=1}^N$ consists of i.i.d. samples drawn from $p_{\text{mix}}$. Let $\mathcal{C}{\subseteq} \mathcal{D}_{\text{train}}$ denote the subset of corrupted samples. We assume $\mathcal{D}_{\text{train}}$ is inaccessible; instead, we only have access to a proxy dataset $\mathcal{D}_{\text{proxy}}{=}\{(x'_j,y'_j)\}_{j=1}^M$ drawn exclusively from the corrupted distribution $p_{\text{corr}}$.
The classification problem is modeled by a conditional distribution $p_\theta(y\mid x)$ over $\mathcal{Y}$, parameterized by $\theta{\in}\Theta{\subseteq}\mathbb{R}^m$, and implemented as a neural network $f{:}\mathcal{X}{\times}\Theta{\to}\mathbb{R}^K$ where $    p_\theta(y\mid x)=\mathrm{softmax}(f(x;\theta))_y.$
The model is trained on $\mathcal{D}_{\text{train}}$ to minimize empirical risk with respect to the training samples. Due to overparameterization, the model can achieve near-zero training error, effectively learning/memorizing the corrupted samples. We denote by $\theta_{\text{mix}}$ the resulting parameters after the training is finished. Once learned/memorized, corrupted samples $\mathcal{C}$ become entangled with clean samples $\mathcal{S} {=} \mathcal{D}_{\text{train}} {\setminus} \mathcal{C}$ in the learned representation, and their influence cannot be trivially removed.

Our goal is to transform $f(\cdot \,;\, \theta_{\text{mix}})$, using only samples in $\mathcal{D}_{\text{proxy}}$, into a corrected model $f(\cdot\,;\,\theta_u)$ that approximates an oracle model $f(\cdot\,;\,\theta_{\text{clean}})$ which is trained only on $\mathcal{S}$.
Let $\mathcal{U}$ be a post-training corrective method that maps the weights and a set of samples to new weights defined as
\begin{equation}
    \mathcal{U}:\Theta\times(\mathcal{X}\times\mathcal{Y})^{M}\to\Theta,
    \qquad
    \theta_u=\mathcal{U}(\theta_{\text{mix}};\mathcal{D}_{\text{proxy}}).
\end{equation}
The correction is considered to be successful if the predictive distributions of the corrected and oracle models
are close in expectation over $p_{\text{mix}}$. Specifically, for a choice of pointwise discrepancy
$d\!\left(f(x;\theta),f(x;\theta')\right)$ (e.g., logit MSE)
the mismatch between the two models is defined as:
\begin{equation}
    \Delta_{\text{mix}}\big(\theta,\theta'\big)
    :=
    \mathbb{E}_{\tilde x\sim p_{\text{mix}}}\!\left[
    d\!\left(f(\tilde x;\theta),\,f(\tilde x;\theta')\right)
    \right].
    \label{eq:cmu_goal_distance}
\end{equation}

The goal is then to obtain $ \theta_u=\mathcal{U}(\theta_{\text{mix}};\mathcal{D}_{\text{proxy}}) $
s.t.  $\Delta_{\text{mix}}\big(\theta_u,\theta_{\text{clean}}\big)\;\le\;\varepsilon,$  for a small tolerance $\varepsilon>0$.

\subsection*{Corruption Kernels}
\label{subsec:corruption_kernels}
Each corruption kernel $\kappa(x',y'{\mid}x,y)$ defines a conditional law over corrupted outputs $(x',y')$ given a clean pair $(x,y){\sim}p_{\text{clean}}$.
For corruptions operating only on labels, inputs remain unchanged, which we express with a Dirac delta $\delta(x'-x)$.

\paragraph{Symmetric Label Noise (SN).}
Each clean label is replaced uniformly by a different class. The kernel is  
\begin{equation}
    \kappa(x',y'\mid x,y)
    =
    \delta(x'-x)\;\frac{\mathbf{1}[y'\neq y]}{K-1}.
    \label{eq:kappa_symmetric}
\end{equation}

\paragraph{Asymmetric Label Noise (AN).}
Let $L{\subset}\mathcal{Y}$ be the subset of classes that are corrupted, and let $g{:}L{\to}\mathcal{Y}$ be a deterministic mapping with $g(y){\neq}y$ for all $y{\in}L$. The kernel is
\begin{equation}
    \kappa(x',y'\mid x,y)=\delta(x'-x)\;\mathbf{1}[\,y\in L\,]\;\mathbf{1}[\,y'=g(y)\,].
    \label{eq:kappa_asymmetric}
\end{equation}

\paragraph{Poison Trigger (PT).}
A trigger operator $\rho{:}\mathcal{X}{\to}\mathcal{X}$ modifies the input, and the label is forced to a target class $y_t{\in}\mathcal{Y}$:
\begin{equation}
    \kappa(x',y'\mid x,y)
    =
    \delta\!\big(x'-\rho(x)\big)\;\mathbf{1}[\,y'=y_t\,].
    \label{eq:kappa_poison}
\end{equation}

\section{Corrective Unlearning in Task Space}
\label{sec:method}
\begin{figure}[t]
    \centering
    \includegraphics[width=0.8\linewidth]{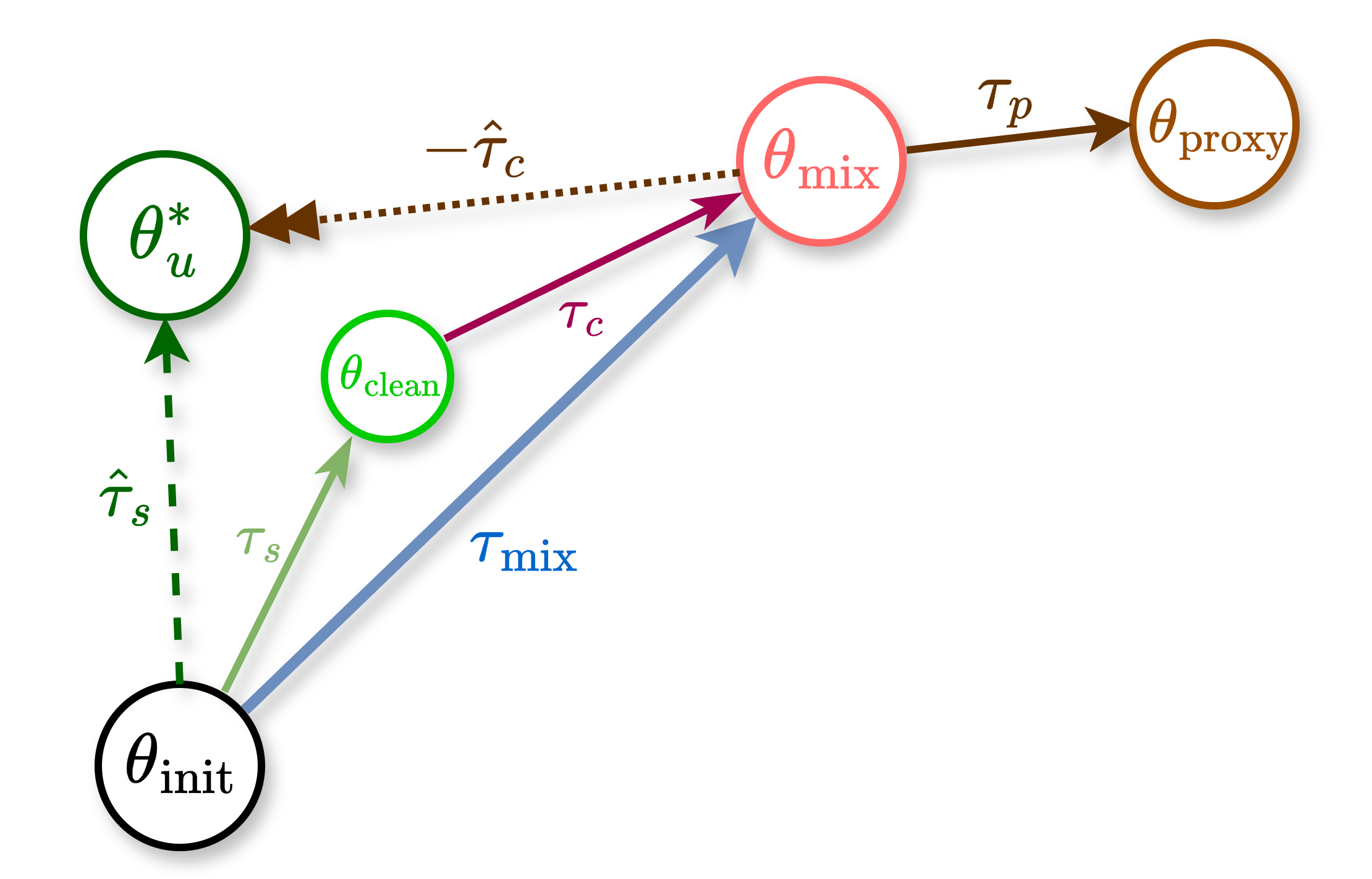}
    \caption{High-level view of CUTS correction procedure in the weight space.} 
    \label{fig:task_arithmetic}
\end{figure}

In this section, we introduce our method. The core steps are summarized in \cref{alg:CUTS}, and an overview of the procedure is shown in \cref{fig:task_arithmetic}.

\begin{algorithm}
\caption{CUTS algorithm for source-free CMU.}
\label{alg:CUTS}
\begin{algorithmic}[1]
    \Statex\textbf{Input:} Corrupted model $\theta_\text{mix}$ trained on partially corrupted training dataset $\mathcal{D}_\text{train} {\sim} p_\text{mix}$; proxy dataset $\mathcal{D}_\text{proxy} {\sim} p_\text{corr}$ containing only corrupted samples.
    \Statex \textbf{Post-training Corrective Method} $\text{CUTS}(\theta_\text{mix}{,} \mathcal{D}_\text{proxy})$:
        \State $\theta_\text{proxy} = \text{finetune}(\theta_\text{mix}, \mathcal{D}_\text{proxy})$
        \State $\tau_p = \theta_\text{proxy} - \theta_\text{mix}$
        \State $\hat\alpha^\ast = \text{alpha\_estimator}(\theta_\text{mix}, \tau_p, \mathcal{D}_\text{proxy})$
        \State $\hat{\theta}^\ast_u = \theta_\text{mix} - \hat\alpha^\ast \, \tau_p$
        \State \Return $\hat{\theta}^\ast_u$
\end{algorithmic}
\end{algorithm}

\subsubsection*{Corruption as a Task}

Considering the mixture distribution $p_\text{mix}$ observed by the learning algorithm, the risk can be decomposed to 
\begin{equation}
    \begin{aligned}
        \mathcal{L}(\theta) &= 
        (1-\eta) \, \mathbb{E}_{(x, y) \sim p_\text{clean}} 
        \big[\ell(f(x \,;\, \theta), y)\big]\\
        &+ \eta \, 
        \mathbb{E}_{(x^\prime, y^\prime) \sim p_\text{corr}} 
        \big[\ell(f(x^\prime \,;\, \theta), y^\prime)\big].
    \end{aligned}
\end{equation}
We view training on $\mathcal{D}_{\text{train}} {\sim} p_\text{mix}$ as a multi-task joint learning consisting of two distinct tasks defined by clean samples $\mathcal{S} {\sim} p_\text{clean}$ and corrupted samples $\mathcal{C} {\sim} p_\text{corr}$. 
$\mathcal{T}_s$ and $\mathcal{T}_c$ denote the tasks defined by $\mathcal{S}$ and $\mathcal{C}$ respectively.

The initialization weights for the model are denoted by $\theta_{\text{init}}$ and the task vector obtained by training on $\mathcal{D}_{\text{train}}$ is $\tau_{\text{mix}}{=}\theta_{\text{mix}}{-}\theta_{\text{init}}$. From the task arithmetic perspective, $\tau_\text{mix}$ can be decomposed to the summation of two directionally distinct task vectors $\tau_s$ and $\tau_c$ associated with $\mathcal{T}_s$ and $\mathcal{T}_c$. That is, $\tau_{\text{mix}}{=} \tau_s{+}\tau_c$. By estimating $\tau_c$, we can recover the clean task vector by $\tau_s{=}\tau_{\text{mix}}{-}\tau_c$
and the clean signal by $\theta_{\text{clean}}{=}\theta_{\text{mix}}{-}\tau_c$. Intuitively, $\tau_c$ captures the incremental update in weight space that intensifies the model’s reliance on corrupted samples and deviation from the original clean signal. In analogy to how task vectors capture skills or behaviors, here the vector $\tau_c$ represents the task of ``learning corruption''. Conversely, going in the negative direction of $\tau_c$, would remove these adverse effects from the model.

\subsubsection*{Source-free Correction}
\label{sec:methodology_source_free_correction_desc}
The original formulation of CMU assumes access to a forget set $\mathcal{D}_f {\subseteq} \mathcal{C} {\subset }\mathcal{D}_{\text{train}}$,
and the goal is to remove the detrimental effect of the corrupted training samples $\mathcal{C}$ from the trained weights $\theta_{\text{mix}}$ using $\mathcal{D}_f$, and potentially to use the retain set $\mathcal{D}_r {=} \mathcal{D}_{\text{train}} {\setminus} \mathcal{D}_f$ for preventing damage to clean signal.
In the source-free setting we assume access to $\mathcal{D}_{\text{proxy}}$ which serves as a proxy or surrogate for $\mathcal{C}$. We use $\mathcal{D}_{\text{proxy}}$ to define a task $\mathcal{T}_p$ and the corresponding task vector $\tau_p$, which is used to estimate the corruption task vector $\tau_c$.


A practical aspect becomes important in the source-free setting. For structured corruptions such as poison triggers, the corruption introduces a consistent and learnable pattern across many samples. As a result, instead of relying on per-sample memorization, the model can generalize the corruption pattern as if it were a true feature of the data distribution. Such a generalizable feature leads to the transferability of the corruption effect between the learned corrupted training samples and unseen samples in a proxy dataset. That is, if a model is affected by poison triggers, its vulnerability can be observed on the proxy dataset, and if we remove the backdoor, the resulting robustness can also be measured there. Consequently, methods designed to mitigate backdoor corruption on identified poisoned samples in the training dataset tend to exhibit similar effectiveness when applied to a proxy set carrying the same trigger \cite{potion}. This, however, contrasts with label noise
, where no shared signal exists between corrupted samples, and the only way for the network to fit the corrupted labels is by memorizing individual samples.
Therefore, per-sample unlearning methods that target memorized training mistakes tend to be ineffective when applied to a proxy set of corrupted samples never seen by the model during training. We show that despite such fundamental differences between corruption types, it is possible to mitigate their effects through CUTS that operates on the underlying weight space mechanism excited by the corruption, rather than on per-sample associations.

\subsubsection*{Estimation of the Corruption Task Vector}
\label{sec:estimation_of_corruption_task_vector}
To estimate $\tau_c$, we fine-tune the partially corrupted model $f(\cdot\,;\,\theta_\text{mix})$ on the proxy dataset $\mathcal{D}_{\text{proxy}}$. This fine-tuning amplifies the corruption patterns already embedded in $\theta_{\text{mix}}$, producing a model $f(\cdot \, ; \, \theta_{\text{proxy}})$ that more distinctly reflects corruption-induced behavior. For instance, under symmetric label noise $f(\cdot \, ; \, \theta_{\text{proxy}})$ behaves nearly as a random classifier, under asymmetric noise it systematically misclassifies corrupted source classes, and under poisoning it becomes more vulnerable to trigger patterns. The weight updates during this process are approximately aligned with the corruption direction defined by $\tau_c$ rather than the clean signal.
We define the corresponding task vector as $\tau_p {=} \theta_{\text{proxy}} {-} \theta_{\text{mix}}$, which captures the corruption-oriented update induced by proxy fine-tuning.  
To correct the model, we move from $\theta_{\text{mix}}$ in the opposite direction of $\tau_p$:
\begin{equation}
     \theta^{\alpha}_u = \theta_{\text{mix}} - \alpha \,\tau_p,
     \label{eq:theta_corrected_estimate}
\end{equation}
where $\alpha \ge 0$ controls the strength of the correction; small $\alpha$ applies a mild adjustment, while larger values more aggressively subtract the corruption component. Next, we define the optimal corrected weights $\theta^{\ast}_u {=} \theta_{\text{mix}} {-} \alpha^{\ast} \tau_p$
where $\alpha^\ast$ is selected to optimize a desired correction objective (e.g., maximizing accuracy on $\mathcal{D}_\text{test}$ or minimizing vulnerability for poison triggers). This choice calibrates the magnitude of the subtraction so that we remove as much of the corruption effect as possible without sacrificing the clean signal. define the estimator $\hat{\tau}_c {=} \alpha^{\ast} \tau_p$.
In other words, $\tau_p$ serves as a (approximately) directionally correct proxy for the corruption task vector $\tau_c$, and $\alpha^\ast$ rescales it to match the strength of the corruption present in $\theta_{\text{mix}}$.

\subsubsection*{Estimating the Optimal Scaling Coefficient} \label{sec:approx_alpha_star}
To eliminate the need for a clean validation dataset, we estimate the optimal scaling coefficient $\alpha^{\ast}$ for the correction operator using only the proxy dataset
$\mathcal{D}_{\text{proxy}}{=}\{(x'_i,y'_i)\}_{i=1}^{M}$. The predicted class for an input $x_i^\prime$ is denoted by $\hat{y}_i^\alpha {=} \argmax_{y} p^\alpha(y\mid x'_i)$ where $p^\alpha$ denotes the output probabilities by $ \theta^{\alpha}_u$.
We search over a finite positive grid $\mathcal{A}$ and denote the resulting estimator by $\hat{\alpha}^{\ast}$ and the corrected weights by this estimator by $\hat{\theta}^\ast_u$  (CUTS).\\
\textbf{Poison triggers.} \label{subsec:approx_alpha_star_poison}
For the poison triggers, we measure the attack success rate (ASR) of $\theta^{\alpha}_{u}$ on $\mathcal{D}_{\text{proxy}}$ and choose the smallest update that reliably suppresses the backdoor
\begin{equation}
\hat{\alpha}^{\ast}
=
\min\Bigl\{
\alpha \in \mathcal{A}
:
\underbrace{
\frac{1}{M}\sum_{i=1}^{M}\mathbf{1}\!\left[\hat{y}_i^\alpha = y_t\right]
}_{\mathrm{ASR}(\alpha)}
\le
\varepsilon
\Bigr\}.
\label{eq:alpha_star_poison}
\end{equation}
with a small threshold $\varepsilon$ (we use $\varepsilon=0.01$). \\
\textbf{Symmetric/asymmetric label noise.} \label{subsec:approx_alpha_star_knn}
In the case of symmetric and asymmetric label noise a targeted behavior which can be evaluated directly on the samples of $\mathcal{D}_{\text{proxy}}$ is not present. Empirically, such noise collapses and overlaps class clusters in the penultimate representation. We therefore use an unsupervised kNN self-agreement criterion on the input images $\{x^\prime_i\}$ in the proxy dataset.

Let $h_i$ be the L2-normalized version of feature outputted by the penultimate embedding of $f(x'_i\, ; \, \theta^{\alpha}_u)$.
Let \(\mathcal{N}_k(i)\) be the indices of the \(k\) nearest neighbors of \(h_i\) (Euclidean distance).
The aggregated self-agreement score is
\begin{equation}
\mathrm{SA}(\alpha)
=
\frac{1}{N}\sum_{i=1}^{N}
\frac{1}{k}\sum_{j\in\mathcal{N}_k(i)}
\mathbf{1}\!\left[\hat{y}_j^\alpha=\hat{y}_i^\alpha\right].
\label{eq:saknndef}
\end{equation}
$\mathrm{SA}(\alpha)$ increases when clusters become tighter and more separated, and decreases when clusters overlap or fragment. To prevent models that predict only a few classes for all samples from achieving a high kNN self-agreement, we penalize the agreement scores by a coverage term which forces the estimator to eliminate falsely corrected models. Considering the number of clusters $|\mathcal{Y}|$ and a coverage rate $\rho\in(0,1]$, which is a hyperparameter depending on the size of proxy dataset, we require only $r=\lceil \rho |\mathcal{Y}|\rceil$ clusters to be sufficiently represented (support at least $k{+}1$ samples). Let $V(\alpha)$ be the number of predicted classes meeting this support; we estimate the optimal scaling coefficient as
\begin{equation}
\hat{\alpha}^{\ast}
\;=\;
\argmax_{\alpha\in\mathcal{A}} \left[
\mathrm{SA}(\alpha)
-
\frac{\max\{0,\, r - V(\alpha)\}}{r}
\right].
\label{eq:alpha_star_knn_cov}
\end{equation}

\subsubsection*{Evaluation Metrics} \label{subsec:evaluation_metrics}
For label corruption, the goal of correction is to maximize generalization. We report Utility (UT\%) as the classification accuracy on the clean test dataset $\mathcal{D}_{\text{test}}$. To quantify how much of the gap between the corrupted model $\theta_{\text{mix}}$ and the oracle $\theta_{\text{clean}}$ is closed by a corrected model $\theta_u$, we use the Recovery Rate (RR\%) defined as
\begin{equation}
\mathrm{RR}(\theta_{u})=\frac{\mathrm{UT}(\theta_u)-\mathrm{UT}(\theta_{\text{mix}})}{\mathrm{UT}(\theta_{\text{clean}})-\mathrm{UT}(\theta_{\text{mix}})}.
\label{eq:recovery_rate}
\end{equation}

For poison triggers, a successful correction must remove the trigger effect while preserving utility. We measure the trigger effect with the Attack Success Rate (ASR\%), the percentage of triggered samples in $\mathcal{C}$ that are classified as the attacker’s target label. Lower is better. We also report an overall score for poison unlearning called Poison Unlearning Score (PUS\%),
\begin{equation}
\text{PUS}(\theta_u) = \mathrm{UT}(\theta_u)\times\bigl(100{-}\mathrm{ASR}(\theta_u)\bigr)/100.
\end{equation}

\section{Experiments}
\label{sec:experiments}

\begin{table*}[ht]
\centering
\scriptsize
\renewcommand{\arraystretch}{1.2}
\setlength{\tabcolsep}{3.5pt}
\caption{Results for CLIP models trained under symmetric (SN) and asymmetric (AN) label noise. Cell values show the utility (UT\%). The rightmost column reports the average recovery rate (RR\%). Best results are shown in bold; second-best results are underlined.}
\label{tab:clip_sym_asym}
\begin{tabular}{
l
ccccc
ccccc
cc
ccccc
cc
|c
}
\toprule
& \multicolumn{5}{c}{MNIST} & \multicolumn{7}{c}{CIFAR10} & \multicolumn{7}{c}{CIFAR100} & \multicolumn{1}{|c}{\multirow{2}{*}{\shortstack{\textbf{Avg.}\\\textbf{RR}}}} \\
\cmidrule(lr){2-6} \cmidrule(lr){7-13} \cmidrule(lr){14-20}
& \multicolumn{5}{c}{SN} & \multicolumn{5}{c}{SN} & \multicolumn{2}{c}{AN} & \multicolumn{5}{c}{SN} & \multicolumn{2}{c}{AN} & \multicolumn{1}{|c}{} \\
\cmidrule(lr){2-6} \cmidrule(lr){7-11} \cmidrule(lr){12-13} \cmidrule(lr){14-18} \cmidrule(lr){19-20}
Model
& 10\% & 20\% & 40\% & 60\% & 80\%
& 10\% & 20\% & 40\% & 60\% & 80\%
& 20\% & 40\%
& 10\% & 20\% & 40\% & 60\% & 80\%
& 20\% & 40\% & \\
\midrule
Mix
& 98.7 & 96.0 & 81.7 & 57.1 & 24.8
& 96.9 & 93.5 & 76.2 & 51.9 & 22.7
& 94.7 & 83.1
& 88.2 & 84.6 & 72.5 & 52.1 & 24.9
& 80.9 & 60.5 & 0.0 \\
Oracle
& 99.7 & 99.7 & 99.7 & 99.6 & 99.5
& 98.6 & 98.7 & 98.5 & 98.4 & 98.0
& 98.7 & 98.4
& 90.2 & 90.2 & 89.8 & 88.9 & 86.9
& 90.2 & 89.4 & 100\\
\cmidrule(lr){1-21}
Mix${-}\tau_{r}$
& 98.7 & 96.1 & 81.8 & 57.1 & 25.2
& 96.8 & 93.5 & 76.2 & 52.0 & 22.7
& 94.7 & 83.1
& 88.2 & 84.5 & 72.4 & 52.0 & 24.7
& 80.9 & 60.6 & -0.2 \\
SAP
& 99.3 & 98.6 & 95.0 & 92.0 & 58.4
& 97.0 & 95.1 & 85.0 & 72.0 & 26.3
& 95.2 & 81.4
& 88.1 & 85.0 & 75.8 & 59.4 & 28.5
& 81.3 & 61.5 & 26.9 \\
CF
& \textbf{99.4} & \underline{98.9} & \underline{97.7} & \underline{96.6} & \textbf{94.0}
& \underline{97.5} & \textbf{97.1} & \textbf{95.6} & \textbf{94.2} & \textbf{90.8}
& \underline{97.2} & \textbf{96.4}
& \underline{87.1} & \underline{84.5} & \underline{82.7} & \textbf{80.0} & \textbf{74.3}
& \underline{86.2} & \textbf{82.8} & \underline{65.0} \\
CUTS
& \underline{99.4} & \textbf{98.9} &\textbf{98.0} & \textbf{96.8} & \underline{84.9}
& \textbf{97.7} & \underline{97.0} & \underline{95.5} & \underline{91.8} & \underline{80.9}
& \textbf{97.4} & \underline{95.5}
& \textbf{88.7} & \textbf{86.8} & \textbf{83.1} & \underline{78.3} & \underline{64.9}
& \textbf{86.3} & \underline{82.2} & \textbf{69.4} \\
\bottomrule
\end{tabular}
\end{table*}

\paragraph{Datasets, corruption injection, and models.}
We evaluate CUTS on three benchmark datasets
MNIST \cite{lecun1998mnist} and 
CIFAR10/100 \cite{krizhevsky2009learning} and on  Clothing1M \cite{clothing1m}, a real-world noisy dataset. 
For experiments involving synthetic corruptions, unless stated otherwise, we hold out 2\% of the original training split as $\mathcal{D}_{\text{proxy}}$, \emph{before} injecting corruption. Then we inject corruption at rate $\eta$ only into the remaining training split to create $\mathcal{D}_{\text{train}}$. All of the samples in $\mathcal{D}_{\text{proxy}}$ are then corrupted with the same corruption kernel as for $\mathcal{D}_{\text{train}}$. For corruption types which do not operate on all classes, we only keep the samples that get corrupted by the corruption kernel in the proxy dataset. For symmetric noise we inject corruption at rates $\eta{\in}\{10\%,20\%,40\%,60\%,80\%\}$ and for asymmetric noise at $\eta{\in}\{20\%,40\%\}$ using dataset-specific mappings $g(y)$ following \cite{asym-noise-def}. Poison trigger is injected at $\eta{\in}\{2\%,10\%,20\%\}$. Following \cite{goel2024correctivemachineunlearning}, we adopt a BadNets-style trigger \cite{badnets} by whitening 3\% of the bottom-right pixels and assigning class 0 to triggered samples.

We use a pre-trained CLIP vision encoder (\texttt{ViT-B/16}) \cite{radford2021learningtransferablevisualmodels} as is common in task arithmetic studies. For MNIST, we additionally include a fully connected network with one hidden layer of dimension 32{,}768 (FC1). For CIFAR10/100, we further consider ResNet18/34/50/101 \cite{resnet} models with both random initialization and ImageNet-pretrained weights. We also perform experiments on DINOv3 (\texttt{ViT-B/16-distilled}) \cite{simeoni2025dinov3} backbone, and vanilla ViT (\texttt{ViT-B/16}) \cite{vit_orig}. We report the results for DINOv3, vanilla ViT, and additional $\eta$ values for ResNets in \cref{sup:sec:extended_results}. \looseness=-1

\paragraph{Baselines and training.} \label{subsec:baselines}
The \textbf{Mix} model ($\theta_{\text{mix}}$) defines the base performance, while the \textbf{Oracle} ($\theta_{\text{clean}}$) is obtained by training $\theta_{\text{init}}$ on the clean set $\mathcal{S}$ and serves as an upper bound for any corrective unlearning method. To demonstrate the effectiveness of our method we employ several baselines. For label corruption, we compare against \textbf{SAP} \cite{kodge2025sap}, the current SOTA CMU method under label noise, which operates on a small set of clean samples (those not memorized under noisy labels during training). For poison triggers, we use \textbf{Potion} \cite{potion}, a SOTA poison unlearning method operating only on a forget set. We also consider a catastrophic forgetting baseline \textbf{CF} \cite{CF}, where we fine-tune $\theta_{\text{mix}}$ on a set of clean samples to induce forgetting of corruption effect. For fair comparison, both SAP and CF use a clean version of $\mathcal{D}_{\text{proxy}}$ (no injected corruption), whereas CUTS and Potion are applied on corrupted $\mathcal{D}_{\text{proxy}}$. Neither SAP nor Potion uses the retain set $\mathcal{D}_r$, so all methods can be evaluated in the same source-free setting. Note that obtaining clean samples can be substantially more costly than assembling corrupted proxies; for instance, under symmetric label noise one can form $\mathcal{D}_{\text{proxy}}$ by taking diverse images and assigning random labels, whereas acquiring their correct labels requires additional annotation effort. Finally, to control for the possibility that improvements arise from generic parameter perturbations rather than targeted negation, we include a random task vector baseline $\textbf{Mix}{-}\boldsymbol{\tau_r}$, where $\tau_r$ matches the layer-wise norm of $\hat{\tau}_c$ as proposed by \cite{ilharco2023editing}.

Across all experiments, we pre-train and fine-tune models using cross-entropy loss and continue training until (near) 100\% training accuracy is reached. This is crucial in the presence of corruption as stopping early would implicitly act as regularization and prevent the full effect of corruption from manifesting in the model.
To obtain the mixed weights $\theta_{\text{mix}}$, we train the model from $\theta_{\text{init}}$ (either pre-trained or randomly initialized, depending on the experiment) on $\mathcal{D}_{\text{train}}$. We then fine-tune $\theta_{\text{mix}}$ on $\mathcal{D}_{\text{proxy}}$ to obtain $\theta_{\text{proxy}}$. The scaling coefficient $\hat{\alpha}^{\ast}$ for CUTS is selected from a grid $\mathcal{A}=\{0.05,\,0.1,\,\ldots,\,4.0\}$. Further training details and hyperparameters are reported in \cref{sup:sec:training_and_hyps}.

\subsection{Results}

\paragraph{Label noise.}
\Cref{tab:clip_sym_asym} reports the results for CLIP models under label noise. Moving along the direction of a random vector $\tau_r$ in the weight space fails to remove the corruption, indicating that the direction of the edit is important. CUTS consistently recovers a large fraction of the gap between Mix and Oracle models and the average recovery is higher than CF (which operates on fully clean labels). Much to our surprise, SAP fails to provide comparable performance with the simple baseline CF (both operating on clean samples). 
We hypothesize that SAP relies heavily on the samples used to project activations being low loss samples, which cannot be identified in the source-free setting. \Cref{tab:resfc_sym_asym} shows the results for randomly initialized networks trained under label noise. We observed that correction is essentially harder (lower RR for all methods) when networks are initialized randomly (before training on $\mathcal{D}_\text{train}$) rather than with pre-trained priors. In \Cref{sec:ablation_analysis}, we analyze the effect of pre-trained priors on the effectiveness of CUTS.


\begin{table}[ht]
\caption{  Utility (UT\%) for randomly initialized models trained under symmetric (SN) and asymmetric (AN) label noise. FC1 is a fully connected network; R18 is ResNet-18. Last column gives average recovery rate (RR\%). Best   in bold; second-best underlined.}
\label{tab:resfc_sym_asym}
\centering
\scriptsize
\renewcommand{\arraystretch}{1}
\setlength{\tabcolsep}{3.5pt}
\begin{tabular}{l ccc ccc ccc | c}
    \toprule
    & \multicolumn{3}{c}{MNIST + FC1}
    & \multicolumn{3}{c}{CIFAR10 + R18}
    & \multicolumn{3}{c}{CIFAR100 + R18} 
    & \multicolumn{1}{|c}{\multirow{2}{*}{\shortstack{\textbf{Avg.}\\\textbf{RR}}}} \\
    \cmidrule(lr){2-4} \cmidrule(lr){5-7} \cmidrule(lr){8-10}
    & \multicolumn{2}{c}{SN} & \multicolumn{1}{c}{AN}
    & \multicolumn{2}{c}{SN} & \multicolumn{1}{c}{AN}
    & \multicolumn{2}{c}{SN} & \multicolumn{1}{c}{AN} & \multicolumn{1}{|c}{} \\
    \cmidrule(lr){2-3} \cmidrule(lr){4-4}
    \cmidrule(lr){5-6} \cmidrule(lr){7-7}
    \cmidrule(lr){8-9} \cmidrule(lr){10-10}
    Model & 20\% & 40\% & 40\%
          & 20\% & 40\% & 40\%
          & 20\% & 40\% & 40\% & \\
    \midrule
    Mix   & 92.9 & 79.7 & 81.9 & 77.9 & 60.0 & 74.5 & 52.6 & 38.3 & 41.5 & 0.0 \\
    Oracle & 98.4 & 98.3 & 98.3 & 88.9 & 87.4 & 88.7 & 66.9 & 63.9 & 63.9 & 100\\
    \cmidrule(lr){1-11}
    $\text{Mix}{-}\tau_r$ & 91.9 & 79.0 & 81.9 & 66.0 & 9.9  & 24.3 & 1.2  & 1.0  & 1.5  & -150.0 \\
    SAP        & \textbf{96.4} & \textbf{93.8} & 86.9 & \underline{81.2} & \underline{67.1} & 75.0 & 53.1 & 40.7 & 42.0 & 27.2 \\
    CF        & 94.9 & 91.1 & \underline{90.9} & 80.0 & \textbf{69.2} & \textbf{80.0} & \underline{53.5} & \underline{41.8} & \textbf{45.8} & \underline{31.3} \\
    CUTS      & \underline{95.0} & \underline{93.6} & \textbf{93.6} & \textbf{81.2} & 65.8 & \underline{79.8} & \textbf{54.4} & \textbf{42.1} & \underline{44.5} & \textbf{34.7} \\
    \bottomrule
\end{tabular}
\end{table}


\paragraph{Poison trigger.}
The results of applying the corrective methods under poison trigger corruption for CLIP are summarized in \cref{tab:clip_dino_pois}. Unlike label noise, backdoor triggers introduce learnable features in the input, which the model learns and associates with the target class. This explains why CF generally fails to remove the trigger; the backdoor feature does not significantly degrade clean signal utility (as indicated by the UT\% of the Mix model), and brief fine-tuning on the clean task cannot undo the feature–target association. In contrast, our method reliably eliminates the trigger while outperforming the SOTA poison unlearning method Potion. Notably, Potion collapses when the corruption rate is high, whereas CUTS shows stronger robustness and can remove the trigger using a proxy set with only one-tenth the size of $\mathcal{C}$.

\begin{table}[ht]
\centering
\caption{Results for CLIP trained under poison trigger corruption. Columns show utility (UT\%) and attack success rate (ASR\%). The right most column shows the average poison unlearning score (PUS). Best results (PUS) are in bold; second-best are underlined.}
\label{tab:clip_dino_pois}
\scriptsize
\renewcommand{\arraystretch}{1.}
\setlength{\tabcolsep}{1.6pt}
\begin{tabular}{lcccccccccccc|c}
\toprule
& \multicolumn{6}{c}{CIFAR10} & \multicolumn{6}{c}{CIFAR100} & \multicolumn{1}{|c}{\multirow{2}{*}{\shortstack{\textbf{Avg.}\\\textbf{PUS}}}} \\
\cmidrule(lr){2-7} \cmidrule(lr){8-13}
& \multicolumn{2}{c}{2\%} & \multicolumn{2}{c}{10\%} & \multicolumn{2}{c}{20\%} & \multicolumn{2}{c}{2\%} & \multicolumn{2}{c}{10\%} & \multicolumn{2}{c}{20\%} & \multicolumn{1}{|c}{\textbf{}}\\
\cmidrule(lr){2-3} \cmidrule(lr){4-5} \cmidrule(lr){6-7} \cmidrule(lr){8-9} \cmidrule(lr){10-11} \cmidrule(lr){12-13}
Model & UT & ASR & UT & ASR & UT & ASR & UT & ASR & UT & ASR & UT & ASR &  \\
\midrule
Mix           & 98.6 & 100    & 98.6 &  100  & 98.5 &  100   & 90.6 & 100  & 90.4 & 100 & 90.4 & 100 & 0.0 \\
Oracle        & 98.5 & 0.0    & 98.5  &  0.0   & 98.5 &   0.0  & 90.9 & 0.0 & 90.6 & 0.0 & 90.2 &  0.0 &  94.5 \\
\cmidrule(lr){1-14}
$\text{Mix}{-}\tau_r$ & 98.6 & 100  & 98.6 &  100 & 98.5 & 100  & 90.7 & 100 & 90.4 & 100 & 90.4 &  100  & 0.0 \\
CF            & 97.8 & 100  & 97.9 & 100  &  97.9 & 100  &  99.6 & 100 & 89.1 & 100 & 88.5 &  100 & 0.0 \\
Potion & \textbf{98.4} & \textbf{0.9} & \underline{10.5} & \underline{0.5} & \underline{10.4} & \underline{0.4} & \underline{87.8} & \underline{1.0} & \underline{73.5} & \underline{0.7} & \underline{49.4} & \underline{1.2} & \underline{54.5} \\
CUTS          & \underline{96.6} & \underline{0.4} & \textbf{96.5} & \textbf{0.9} & \textbf{97.6} & \textbf{1.0} & \textbf{88.8} & \textbf{1.7} & \textbf{86.4} & \textbf{1.6} & \textbf{82.1}& \textbf{1.0} & \textbf{90.3} \\
\bottomrule
\end{tabular}
\end{table}

\paragraph{Real-world dataset.}
We tested CUTS on Clothing1M \citep{clothing1m}, a real-world noisy dataset. For the proxy dataset, we randomly sampled 1000 images from the validation set and corrupted their labels with symmetric noise. Although the original dataset may involve a more complex noise transition matrix, we assume it is unknown and use the symmetric noise (one can use noise transition matrix estimation methods for better results). We trained a classifier with DINOv3 backbone to reach near perfect training accuracy (99.89\%). The results of correction are summarized in \cref{tab:dino_clothing1m}; CUTS recovers an average of 2.84\% utility over Mix.


\begin{table}
\centering
\scriptsize
\renewcommand{\arraystretch}{1.2}
\setlength{\tabcolsep}{4pt}
\caption{%
Results on DINOv3 trained on Clothing1M, averaged over 4 runs with different proxy draws and noise seeds.
}
\begin{tabular}{lcc}
\toprule
Model  & {UT} & Improvement \\
\midrule
Mix & 66.42  & - \\
\cmidrule(lr){1-3}
$\text{Mix}{-}\tau_r$ & 66.47 $\pm$ 0.03 & 0.05 \\
CF & 71.64 $\pm$ 0.26 & 5.22\\
\cmidrule(lr){1-3}
CUTS & 69.26 $\pm$ 0.85  & 2.84\\
\bottomrule
\end{tabular}
\label{tab:dino_clothing1m}
\end{table}

\subsection{Ablation and Analysis}
This section presents the main ablation studies; additional analyses including the effect of the
size of $\mathcal{D}_\text{proxy}$ and the cosine similarity between
task vectors are reported in \cref{sup:sec:extended_analysis}.
\label{sec:ablation_analysis}

\begin{figure}[ht]
    \centering
    \includegraphics[width=0.8\linewidth]{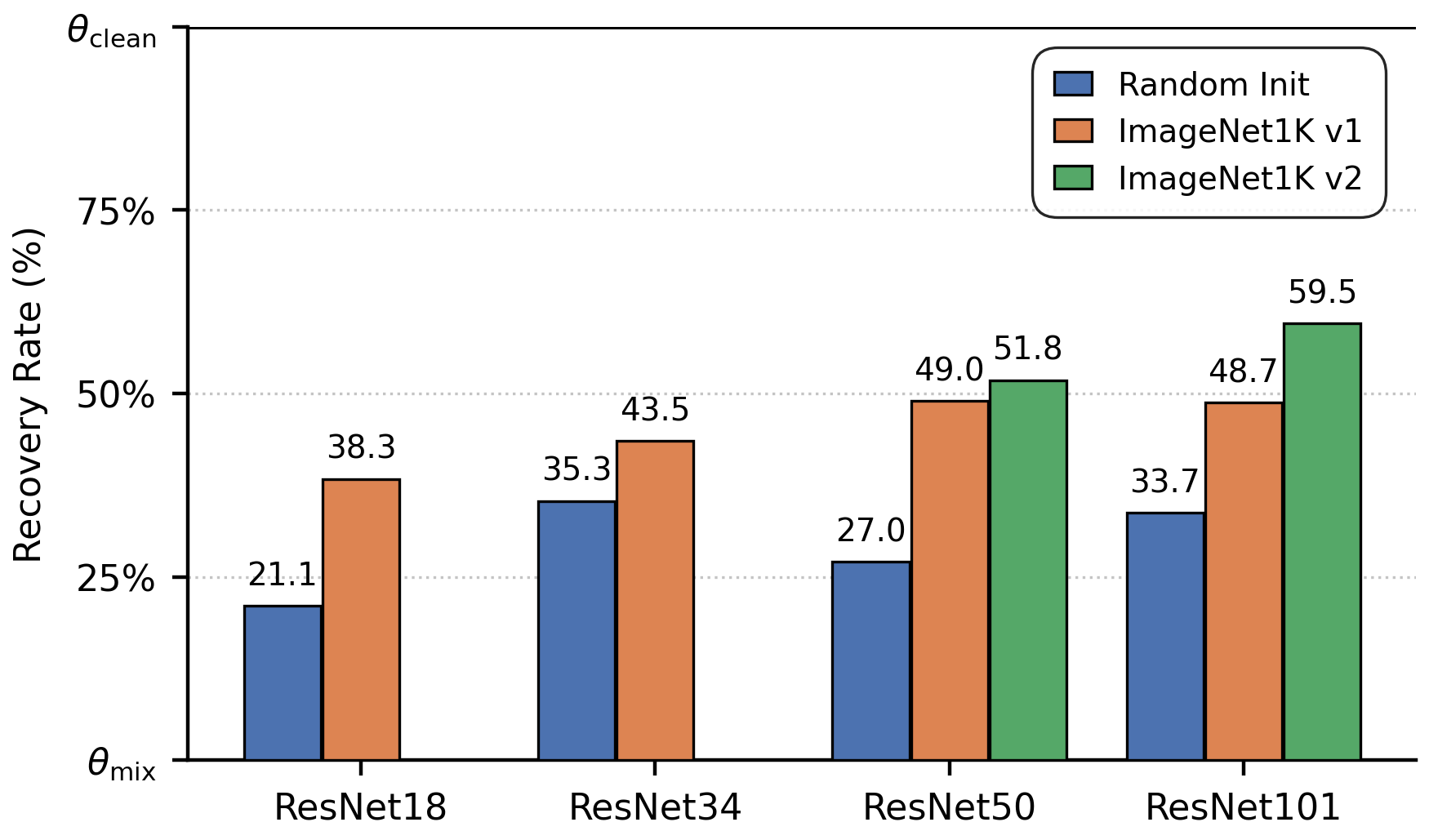}
    
    \caption{RR\% of CUTS on ResNet models with different initializations on CIFAR10 with symmetric label noise at $\eta{=}40\%$.}
    \label{fig:pretrained_priors}
\end{figure}

\paragraph{Pre-trained priors.}
Task arithmetic is most reliable when the network exhibits weight disentanglement and this disentanglement is an emergent property of large pre-training rather than a generic property of neural networks \cite{ortiz-jimenez2023task}. \Cref{fig:pretrained_priors} illustrates the effect of pre-trained priors on the effectiveness of CUTS for ResNet models initialized randomly and with ImageNet1K weights. We observe that better pre-training leads to better disentanglement of the directions associated with corruption and clean tasks in the weight space, which directly impact the performance of CUTS.

\begin{figure}[ht]
    
    \centering
    \includegraphics[width=\linewidth]{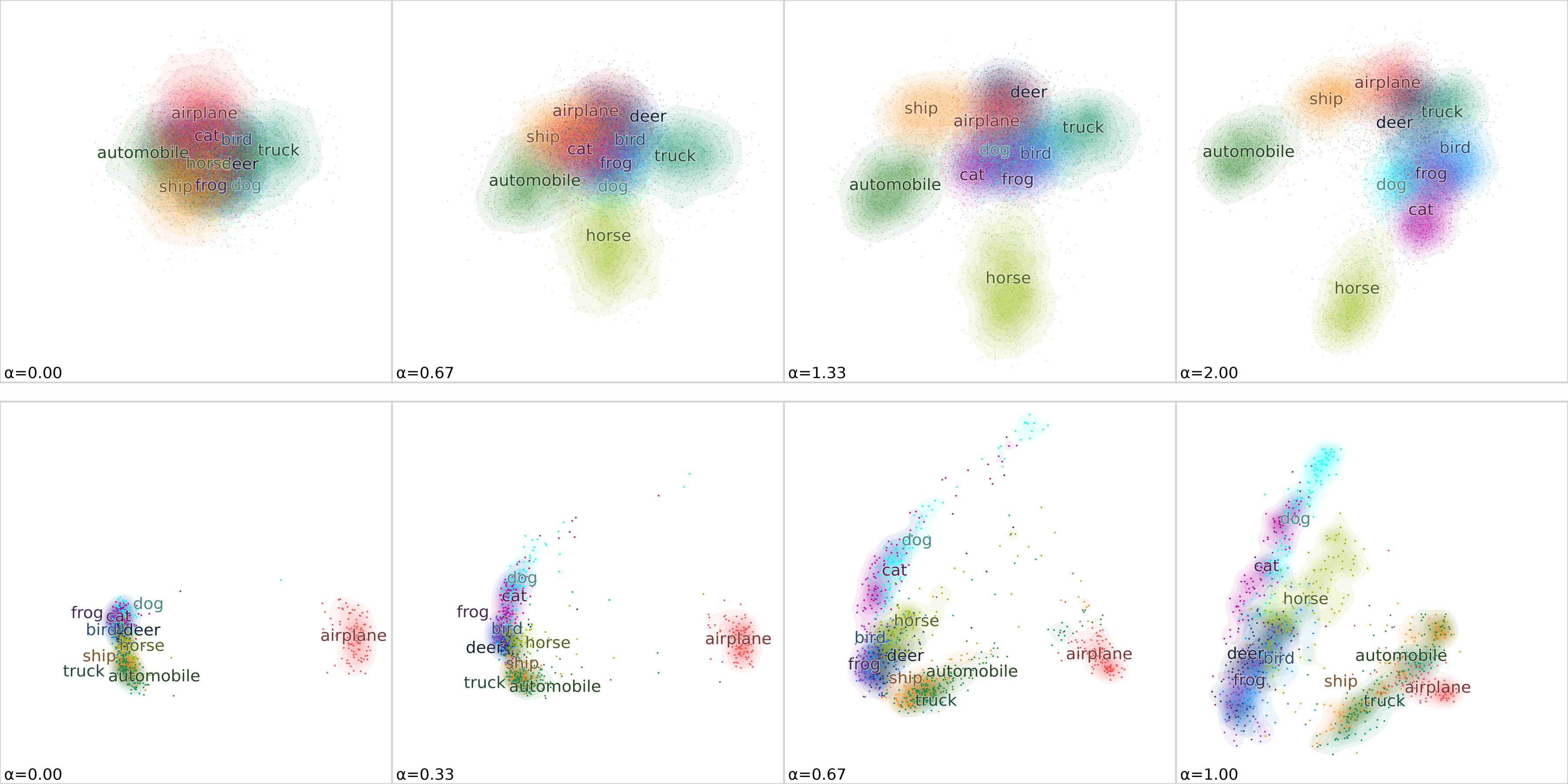}
    \caption{PCA of the penultimate features of CLIP for $\theta_\text{mix}- \alpha {\tau}_p $ , varying $\alpha \in \{ 0,  \hat{\alpha}^*/3, 2 \hat{\alpha}^*/3, \hat{\alpha}^*\}$. Top: CIFAR10 under symmetric label noise at $\eta{=}60\%$. Bottom: CIFAR10 under poison trigger corruption at $\eta{=}2\%$ with target class \textit{airplane}.}
    \label{fig:pca_evol_clip}
\end{figure}

\paragraph{PCA trajectory.}
In order to analyze how CUTS alters the corrupted model, we inspect the embedding space by applying PCA to the penultimate features of $\theta_\text{mix}- \alpha {\tau}_p $, varying $\alpha \in \{ 0,  \hat{\alpha}^*/3, 2 \hat{\alpha}^*/3, \hat{\alpha}^*\} $. Results are shown in \cref{fig:pca_evol_clip}. For label noise, class embeddings are initially collapsed/overlapping due to the high fraction of noisy labels, and increasing the correction strength (i.e. gradually increasing $\alpha$) progressively separates the class clusters and partially restores the decision boundaries. For poison trigger, the triggered inputs are mapped to a compact cluster that is distinct from clean samples of the
same target class. As $\alpha$ increases, this triggered
cluster dissolves and the samples migrate into their true class clusters.

\begin{figure}[ht]
\centering
    \begin{subfigure}[t]{0.32\linewidth}
      \centering
      \includegraphics[width=\linewidth]{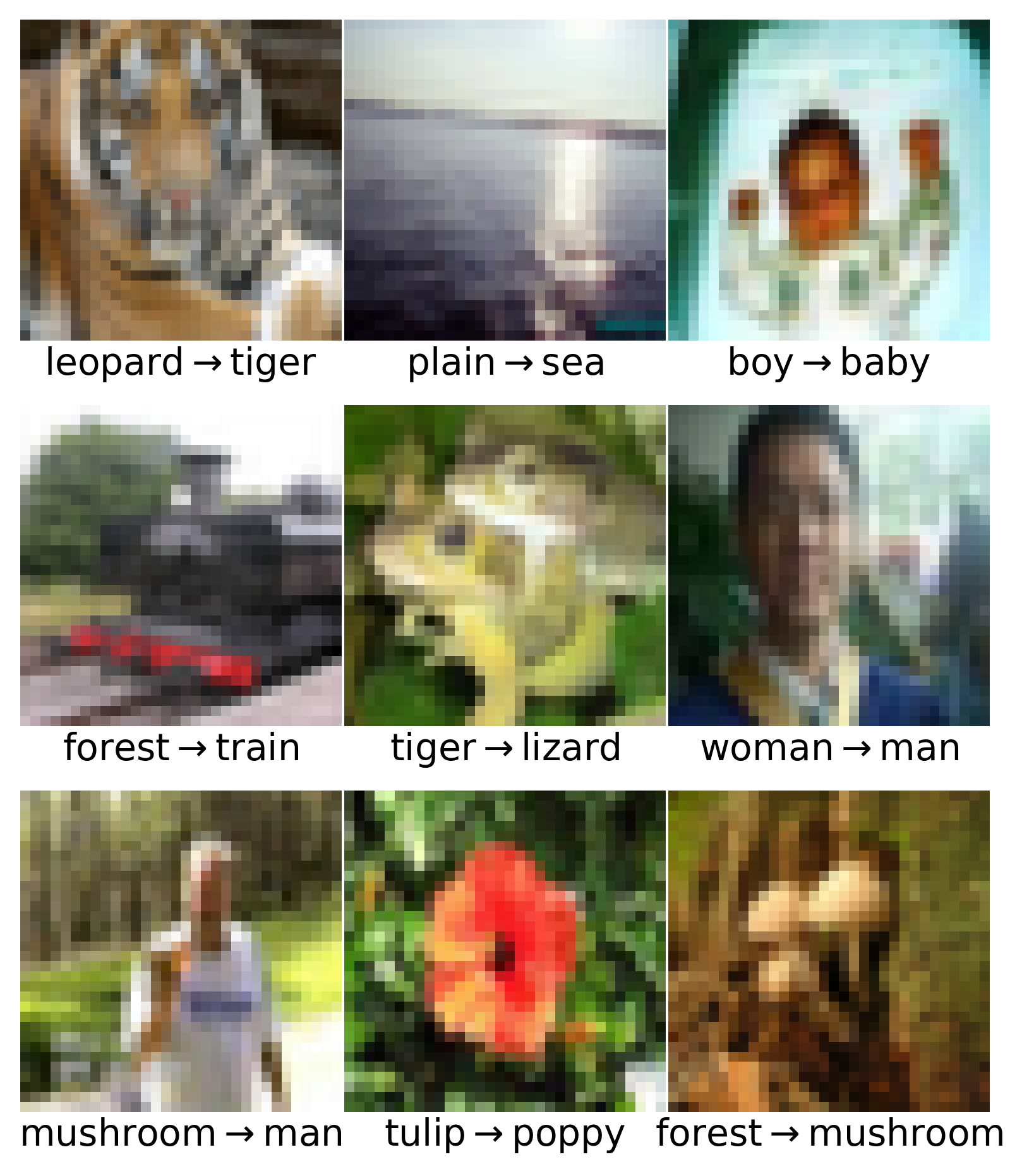}
      \caption{CIFAR100}
    \end{subfigure}\hfill
    \begin{subfigure}[t]{0.32\linewidth}
      \centering
      \includegraphics[width=\linewidth]{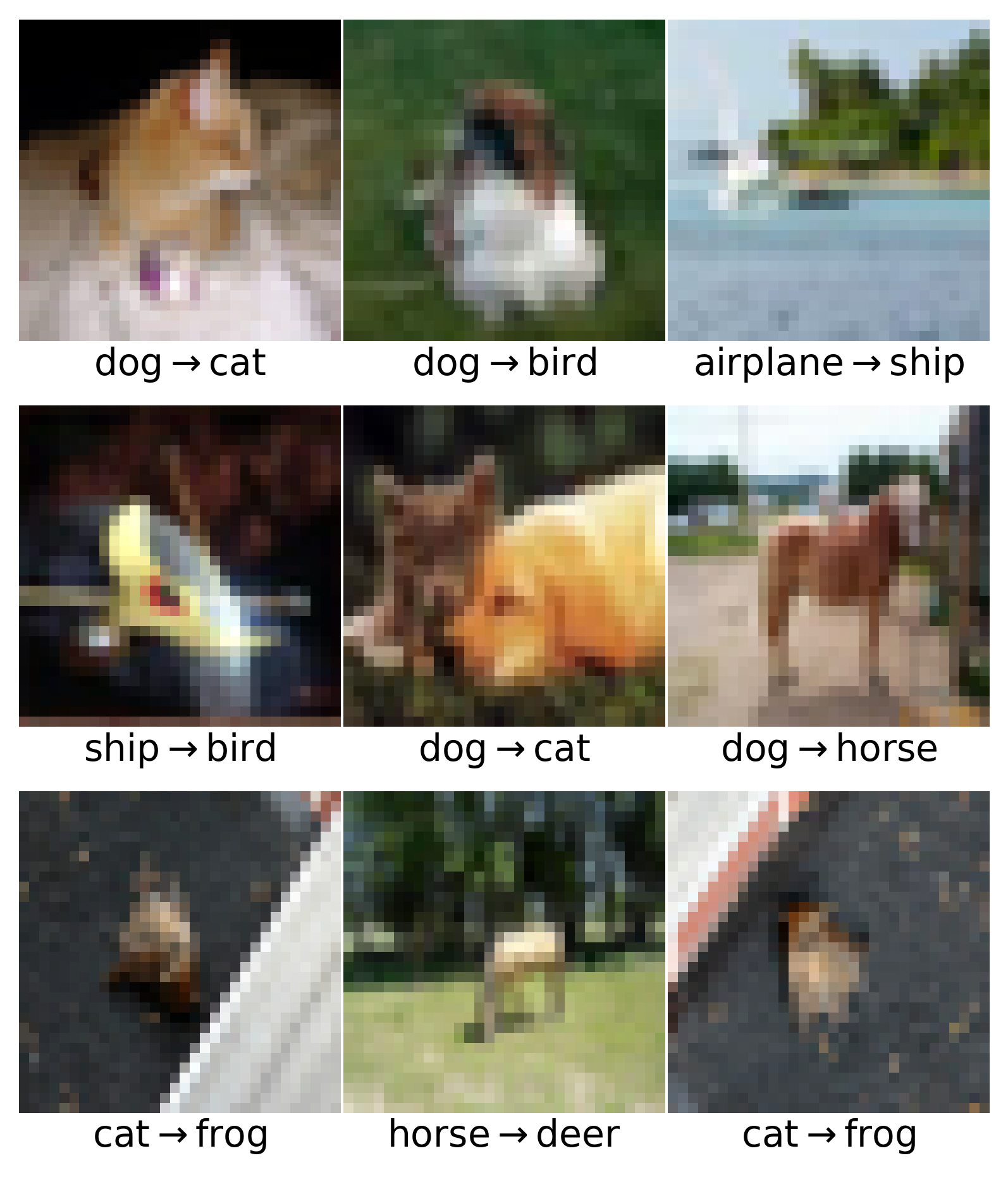}
      \caption{CIFAR10}
    \end{subfigure}\hfill
    \begin{subfigure}[t]{0.32\linewidth}
      \centering
      \includegraphics[width=\linewidth]{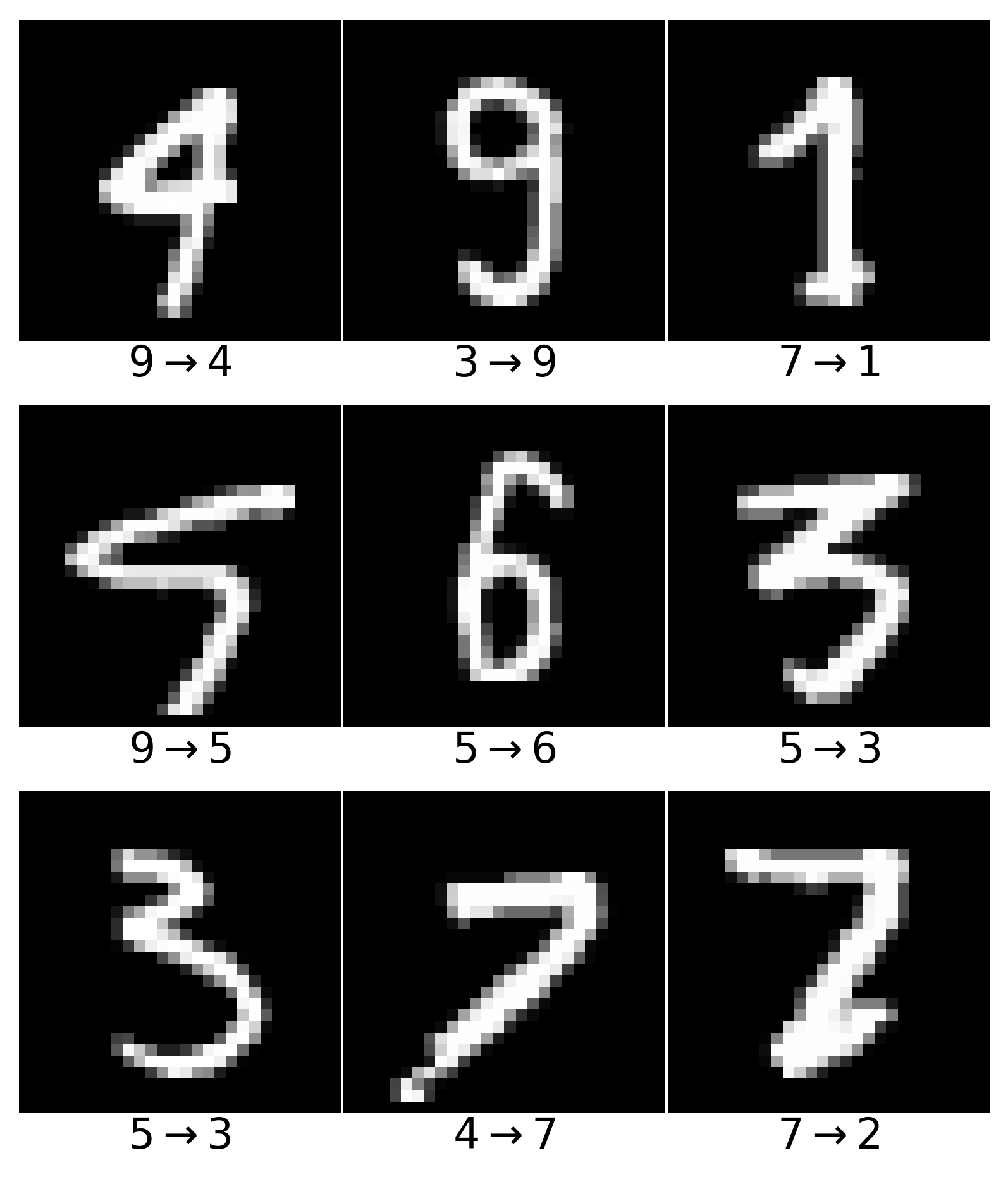}
      \caption{MNIST}
    \end{subfigure}
\caption{Clean training samples that flip first when increasing $\alpha$ along $-\tau_p$ for CLIP trained under symmetric label noise at $\eta{=}10\%$. Captions show \emph{ground truth $\rightarrow$ predicted}.}
\label{fig:forgotten_samples_sym_noise}
\end{figure}

\paragraph{Correcting ``clean'' samples in the original dataset.}
As correction progresses along the correction ray $-\tau_p$ by increasing $\alpha$, in addition to forgetting samples from $\mathcal{C}$, the model also gradually changes its prediction on a number of samples in $\mathcal{S}$. By inspecting these samples (as shown in \cref{fig:forgotten_samples_sym_noise}), we observed that many of the earliest ``clean'' flips are either mislabeled and ambiguous or  atypical and low-support instances that sit near the decision boundaries.

\begin{figure}[ht]
    \centering
    \includegraphics[width=\linewidth]{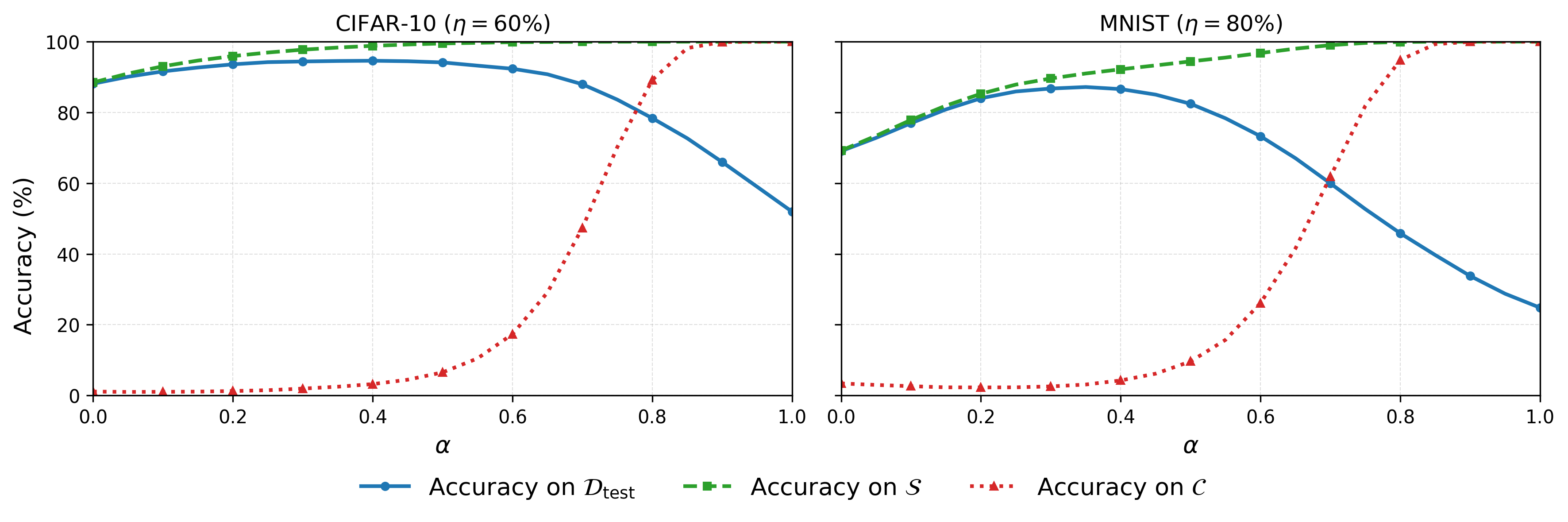}
    \caption{Accuracy of the interpolated CLIP models along $\tau_\text{mix}$ on test dataset and clean and noisy splits ($\mathcal{S}$ and $\mathcal{C}$) of the training datasets of CIFAR-10 and MNIST under symmetric label noise.}
    \label{fig:mix_interpolation}
\end{figure}

\paragraph{Interpolation along $\boldsymbol{\tau}_\text{mix}$ and post-training early stopping.}
Studies have shown that during empirical risk minimization, neural networks tend to learn simple and frequent patterns first, which usually correspond to clean and easy examples, and only later in the training, begin to memorize rare, hard, and noisy samples \cite{arpit2017closerlookmemorizationdeep, nakkiran2019sgdneuralnetworkslearns, maini2023neuralnetworkmemorizationlocalized}. Despite this ordering, the training trajectory in weight space is often highly nonlinear for randomly initialized models. For pre-trained models, however, fine-tuning tends to move in a lower dimensional and more linearly connected region of the loss landscape \cite{garipov2018loss, pmlr-v80-draxler18a, ilharco2023editing, pmlr-v162-wortsman22a}. We tested whether the approximately linear behavior often seen during fine-tuning also appears when the training data are partially corrupted, and whether it aligns with the ``simple to noisy'' learning order described. To do so, we interpolated along $\tau_\text{mix}$ and evaluated each model on $\mathcal{D}_\text{test}$, $\mathcal{S}$, and $\mathcal{C}$. As shown in \cref{fig:mix_interpolation}, as the model moves from $\theta_\text{init}$ to $\theta_\text{mix}$ along $\tau_\text{mix}$, accuracy on $\mathcal{S}$ increases first while accuracy on $\mathcal{C}$ stays low, which leads to a rise in test accuracy. After the clean samples are largely fit, the model begins to fit noisy samples and test accuracy begins to drop. These results are consistent with an approximately linear fine-tuning path for pre-trained models even under substantial label noise and with the known training time ordering where clean examples are learned before noisy ones. Practically, this unlocks a simple post-training early stopping possibility. One can sweep the interpolation coefficient along $\tau_{\text{mix}}$ and pick the point that maximizes validation accuracy, which can recover the best generalization without rerunning training (in case $\theta_\text{init}$ are available).

\section{Conclusion}
\label{sec:conclusion}
In this work, we formalized source-free CMU, which tackles a realistic problem where training data is not available for correction, and introduced CUTS which applies CMU using a small set of corrupted samples which do not need to be from the training set of the model. Particularly, we treat corruption as a task in the task space and operate on task directions in the weight space instead of per-sample associations. We also showed the effect of pre-trained priors on disentanglement of the clean and corruption tasks in the weight space. CUTS gains on randomly initialized networks are smaller which is in parallel with previous observations in task arithmetic works \cite{ortiz-jimenez2023task}. One possible extension of our work can be a zero-shot version of CMU, where the proxy set is generated directly from the trained models using data-free inversion methods \cite{yin2020dreaming, 9010308}. Scaling coefficient $\alpha$ can also be tuned layer-wise instead of a single global coefficient. 
\newpage
{
    \small
    \bibliographystyle{ieeenat_fullname}
    \bibliography{main}
}

\appendix
\clearpage
\setcounter{page}{1}
\maketitlesupplementary

\begin{table*}[ht]
\centering
\scriptsize
\renewcommand{\arraystretch}{1.2}
\setlength{\tabcolsep}{3.5pt}
\caption{Full results for FC1 (for MNIST) and randomly initialized ResNet18 models trained under symmetric (SN) and asymmetric (AN) label noise. Cell values show the utility (UT\%). The rightmost column reports the average recovery rate (RR\%). Best results are shown in bold; second-best results are underlined.}
\label{tab:full_resnet_fc_noise_results}
\begin{tabular}{lccccccc c ccccccc c ccccccc|c}
\toprule
& \multicolumn{7}{c}{MNIST+FC1} & & \multicolumn{7}{c}{CIFAR10+ResNet18} & & \multicolumn{7}{c}{CIFAR100+ResNet18} & \multicolumn{1}{|c}{\multirow{2}{*}{\shortstack{\textbf{Avg.}\\\textbf{RR}}}} \\
\cmidrule(lr){2-8} \cmidrule(lr){10-16} \cmidrule(lr){18-24}
& \multicolumn{5}{c}{SN} & \multicolumn{2}{c}{AN} & & \multicolumn{5}{c}{SN} & \multicolumn{2}{c}{AN} & & \multicolumn{5}{c}{SN} & \multicolumn{2}{c}{AN} & \multicolumn{1}{|c}{}\\
\cmidrule(lr){2-6} \cmidrule(lr){7-8} \cmidrule(lr){10-14} \cmidrule(lr){15-16} \cmidrule(lr){18-22} \cmidrule(lr){23-24}
Model & 10\% & 20\% & 40\% & 60\% & 80\% & 20\% & 40\% & & 10\% & 20\% & 40\% & 60\% & 80\% & 20\% & 40\% & & 10\% & 20\% & 40\% & 60\% & 80\% & 20\% & 40\% & \\
\midrule
Mix & 96.5 & 92.9 & 79.7 & 56.2 & 26.3 & 94.2 & 81.9 & & 83.4 & 77.9 & 60.0 & 38.8 & 16.6 & 84.2 & 74.5 & & 61.0 & 52.6 & 38.3 & 20.5 & 6.7 & 56.7 & 41.5 & 0.0 \\
Oracle & 98.5 & 98.4 & 98.3 & 97.9 & 97.1 & 98.4 & 98.3 & & 89.7 & 88.9 & 87.4 & 85.0 & 79.5 & 89.7 & 88.7 & & 68.1 & 66.9 & 63.9 & 57.5 & 45.4 & 67.6 & 63.9 & 100 \\
\cmidrule(lr){1-25}
$\text{Mix}{-}\tau_{r}$ & 95.6 & 91.9 & 79.0 & 55.5 & 26.2 & 94.2 & 81.9 & & 75.9 & 66.0 & 9.9 & 10.1 & 10.2 & 79.9 & 57.3 & & 44.2 & 1.2 & 1.0 & 0.9 & 2.1 & 33.7 & 1.5 & -92.64 \\
SAP & \textbf{97.1} & \textbf{96.4} & \textbf{93.8} & \textbf{91.0} & 69.1  & \underline{96.9} & 86.9 & & \underline{83.8} & \textbf{81.2} & \underline{67.1} & 48.4 & \underline{19.9} & \textbf{85.4} & 75.0 & & \textbf{61.4} & 53.1 & 40.7 & \underline{24.0} & \underline{8.7} & \underline{57.1} & 42.0 & \underline{26.56}  \\
CF & 96.4 & 94.9 & 91.1 & 87.5 & \textbf{80.1} & 95.0 & \underline{90.9} & & 83.6 & \underline{80.0} & \textbf{69.2} & \textbf{57.2} & \textbf{32.0} & \underline{84.1} & \textbf{80.0} & & \underline{60.9} & \underline{53.5} & \underline{41.8} & \textbf{26.1} & \textbf{11.4} & 57.0 &  \textbf{45.8} & 25.84 \\
CUTS & \underline{96.5} & \underline{95.0} & \underline{93.6} & \underline{88.4} & \underline{71.4} & \textbf{97.3} & \textbf{93.6} & & \textbf{84.3} & \textbf{81.2} & \underline{65.8} & \underline{52.7} & 19.5 & 83.9 & \underline{79.8} & & 60.2 & \textbf{54.4} & \textbf{42.1} & 23.3 & 7.9 & \textbf{57.3} & \underline{44.5} & \textbf{27.46} \\
\bottomrule
\end{tabular}
\end{table*}

\section{Training and Hyperparameter Details}
\label{sup:sec:training_and_hyps}
\subsection{Model Details}
\label{sup:subsec:training_details_models}

We use a pre-trained CLIP vision encoder (\texttt{ViT-B/16}) with original OpenAI pre-trained weights. Following standard practice, we render a number of predefined templates with class names in each dataset, encode them with the CLIP text encoder to build per-class prototypes, average templates per class, and freeze the resulting prototype linear head. The text encoder is then removed from the model, and we fine-tune only the vision encoder. Task vectors are then constructed using the parameters of the vision encoder (excluding the frozen classification head). Randomly initialized networks use PyTorch’s default initialization algorithm, and for pre-trained ResNets and vanilla ViT (\texttt{ViT-B/16}), we use \texttt{ImageNet1Kv1} weights from \texttt{torchvision}. Furthermore, for the analysis of pre-trained priors with ResNet models, we also use \texttt{ImageNet1Kv2} weights when available (ResNet50/101), which are slightly more powerful than \texttt{ImageNet1Kv1}. As for DINOv3, we use the version \texttt{facebook/dinov3-vitb16-pretrain-lvd1689m} from HuggingFace.

We note that, when computing task vectors for ResNet models, we exclude all BatchNorm (BN) parameters and buffers (i.e., the affine weights $(\gamma,\beta)$ and the running mean/variance) and retain their values from $\theta_{\text{mix}}$. This is because fine-tuning on a fully corrupted proxy dataset induces large shifts in BN running statistics and if these BN deltas are included and later negated from $\theta_{\text{mix}}$, they overwrite partially calibrated statistics, destabilize activations, and severely degrade performance. Because we cannot reliably recalibrate BN without clean data in our source-free setting, we treat BN as fixed and compute task vectors over the remaining parameters.

\subsection{CUTS Training Procedure}
\label{sup:subsec:training_details_CUTS}

Across all experiments, we pre-train and fine-tune using the cross-entropy loss and the AdamW optimizer \cite{loshchilov2018decoupled, pytorch} with default hyperparameters, except that we tune the learning rate. Our primary learning rate schedule is cosine annealing with a linear warmup \cite{cosineannealingLR}. We extend this schedule with a \emph{hold-steps} hyperparameter that keeps the peak learning rate constant for a fixed number of steps before decay begins; this helps the model fully fit the corrupted samples and prevents premature annealing. For some experiments with ResNets and FC1, we instead use a MultiStep schedule for fine-tuning. Full training details, including the number of iterations, learning rate, schedule type, and corresponding hyperparameters, are reported in \cref{tab:training_hyperparams}.

For input preprocessing, we use the default transforms provided with CLIP and DINOv3. For FC1, we apply no augmentations and only normalize images using the dataset-specific mean and standard deviation. For ResNets and vanilla ViT, we use \texttt{RandomCrop} with 4 pixels of padding and normalize with dataset-specific statistics when the network is randomly initialized and ImageNet statistics when initialized from \texttt{ImageNet1Kv1} or \texttt{ImageNet1Kv2} pre-trained weights.

Additionally, for symmetric label noise experiments, CUTS results are obtained by negating the average of two proxy task vectors, $\tau_{p}^1$ and $\tau_{p}^2$, each computed from a proxy dataset $\mathcal{D}_\text{proxy}^i$ with identical input images but independently corrupted labels (different random seeds). This is approximately equivalent to negating each vector separately and then averaging the resulting weights; it leaves the overall conclusions unchanged while providing a slightly lower-variance performance estimate.

For the scaling coefficient estimator $\hat\alpha^\ast$ in the case of label noise, we calculate the self-agreement score with
\begin{equation*}
    k = \frac{|\mathcal{D}_\text{proxy}|}{2\,|\mathcal{Y}|}.
\end{equation*}
We observed that the algorithm is robust to the choice of $k$ as long as it is set to be larger than 5 and smaller than $|\mathcal{D}_\text{proxy}|/|\mathcal{Y}|$. For the coverage rate $\rho$, we use $\rho{=}1$ for all datasets (including Clothing1M) except for CIFAR100 where we use $\rho{=}0.95$, as the proxy dataset only contains 10 samples for each class and the corrected model outputted by CUTS, might not predict those specific samples correctly if the samples are hard or boundary instances.
\subsection{Baselines}
\label{sup:subsec:training_details_baselines}
\paragraph{Catastrophic forgetting.} The hyperparameters used for fine-tuning the Mix model on the $\mathcal{D}_\text{proxy}$ dataset without corruption injection are reported in \cref{tab:training_hyperparams}. We fine-tune the models to reach near 100\% training accuracy.

\paragraph{SAP.} For SAP, we use the input images in $\mathcal{D}_\text{proxy}$ for projecting the activations. SAP has a hyperparameter $\alpha$ which the authors state works best when set to $10000$. However, we observed that this value would not lead to acceptable results for all noise rates $\eta$. Hence, we used the clean version of $\mathcal{D}_\text{proxy}$ as a validation set to tune $\alpha$ and chose the value from the grid $\mathcal{G} = \{1000,\allowbreak 5000,\allowbreak 10000,\allowbreak 20000,\allowbreak 30000,\allowbreak 40000,\allowbreak 50000,\allowbreak 70000,\allowbreak 100000,\allowbreak 200000,\allowbreak 300000,\allowbreak 400000,\allowbreak 500000,\allowbreak 1000000\}$.

\paragraph{Potion.} Potion only uses one hyperparameter which is the fraction of the dataset used for adaptive percentile calculation. We set this hyperparameter to be $0.01$ since the implementation adaptively increases this value until it can bring ASR on $\mathcal{D}_\text{proxy}$ below $1\%$. Furthermore, we set the maximum number of tries to 500, so that the algorithm terminates with the specified condition in all experiments.

\subsection{Reproducibility}
\label{sup:subsec:training_details_reproducibility}
All experiments were conducted inside a PyTorch NGC Container with version \texttt{pytorch:25.05-py3}, which provides a fixed and portable software environment. The container includes \texttt{Python 3.12.3}, \texttt{PyTorch 2.8.0.nv25.05}, and the \texttt{CUDA 12.9} runtime. Experiments were conducted on a single \texttt{NVIDIA RTX~A6000} GPU (48\,GB VRAM), while those on the Clothing1M dataset utilized two GPUs. To guarantee deterministic results, a global random seed of~11 was fixed and propagated across all libraries (\texttt{Python}, \texttt{NumPy}, and \texttt{PyTorch}) and devices.


\section{Extended Results}
\label{sup:sec:extended_results}
\subsection*{Label Corruption}
\label{sup:subsec:extended_results_label_noise}
\Cref{tab:full_resnet_fc_noise_results} provides a complete overview of the performance of the correction methods on randomly initialized FC1 and ResNet18 models across all noise rates $\eta$. It serves as the extended version of \cref{tab:resfc_sym_asym}.

Furthermore, \cref{tab:vit_label_noise} summarizes the behavior of the vanilla ViT model under label noise. We also include in \cref{tab:dino_label_noise} the corresponding evaluation of CUTS on classifiers with a DINOv3 backbone (only at $\eta{=}40\%$). Unlike for the CLIP architecture, for which we implemented SAP, the SAP baseline was not implemented for this backbone.

As noted before, CUTS performs particularly well when the initial parameters $\theta_{\text{init}}$ correspond to strong pre-trained weights. Its recovery rate does decrease for randomly initialized networks; however, a similar degradation is observed for all competing baselines. Based on the average recovery rates reported in the rightmost columns of \cref{tab:full_resnet_fc_noise_results,tab:vit_label_noise,tab:dino_label_noise}, CUTS achieves the highest average performance in every architecture–initialization setting, despite not having access to clean labels.

\begin{table*}[ht]
\centering
\scriptsize
\renewcommand{\arraystretch}{1.2}
\setlength{\tabcolsep}{3.5pt}
\caption{Results for ViT-B/16 model initialized with \texttt{ImageNet1Kv1} weights trained under symmetric (SN) and asymmetric (AN) label noise. Cell values show the utility (UT\%). The rightmost column reports the average recovery rate (RR\%). Best results are shown in bold; second-best results are underlined.}
\label{tab:vit_label_noise}
\begin{tabular}{l ccccccc c ccccccc | c}
\toprule
& \multicolumn{7}{c}{CIFAR10} & & \multicolumn{7}{c}{CIFAR100} & \multicolumn{1}{|c}{\multirow{2}{*}{\shortstack{\textbf{Avg.}\\\textbf{RR}}}} \\
\cmidrule(lr){2-8} \cmidrule(lr){10-16}
& \multicolumn{5}{c}{SN} & \multicolumn{2}{c}{AN} & & \multicolumn{5}{c}{SN} & \multicolumn{2}{c}{AN} & \multicolumn{1}{|c}{}\\
\cmidrule(lr){2-6} \cmidrule(lr){7-8} \cmidrule(lr){10-14} \cmidrule(lr){15-16}
Model & 10\% & 20\% & 40\% & 60\% & 80\% & 20\% & 40\% & & 10\% & 20\% & 40\% & 60\% & 80\% & 20\% & 40\% & \\
\midrule
Mix    & 93.6 & 90.5 & 72.7 & 53.7 & 21.5 & 90.6 & 78.5 & & 78.4 & 74.2 & 61.3 & 42.2 & 19.1 & 72.0 & 52.7 & 0.0 \\
Oracle & 96.1 & 95.9 & 93.7 & 94.2 & 93.2 & 95.6 & 95.4 & & 82.3 & 82.1 & 80.0 & 78.5 & 73.4 & 81.5 & 80.1 & 100 \\
\cmidrule(lr){1-16}
$\text{Mix}{-}\tau_{r}$ & 93.5 & 90.5 & 73.5 & 54.0 & 21.6 & 90.6 & 78.5 & & 78.3 & 74.2 & 61.4 & 42.4 & 19.1 & 72.1 & 52.7 & 0.22 \\
SAP    & \textbf{93.6} & \textbf{91.9} & 75.4 & 64.1 & 24.3 & 90.8 & 78.6 & & 78.0 & 74.5 & 63.1 & 43.7 & 19.0 & 72.5 & 52.6 & 6.35 \\
CF     & \underline{93.5} & 91.6 & \underline{79.7} & \underline{75.0} & \textbf{49.1} & \underline{91.7} & \underline{87.9} & & \textbf{78.7} & \underline{75.5} & \textbf{66.9} & \textbf{53.6} & \textbf{33.0} & \underline{74.0} & \underline{60.6} & \underline{27.29} \\
CUTS   & 93.4 & \underline{91.8} & \textbf{81.6} & \textbf{77.7} & \underline{41.1} & \textbf{92.8} & \textbf{88.7} & & \underline{78.4} & \textbf{76.0} & \underline{66.4} & \underline{52.8} & \underline{25.4} & \textbf{74.8} & \textbf{62.3} & \textbf{29.12} \\
\bottomrule
\end{tabular}
\end{table*}

\begin{table}[ht]
\centering
\scriptsize
\renewcommand{\arraystretch}{1.2}
\setlength{\tabcolsep}{3.5pt}
\caption{Results for classifiers with DINOv3 backbone trained under symmetric (SN) and asymmetric (AN) label noise at $\eta{=}40\%$. Cell values show the utility (UT\%). The rightmost column reports the average recovery rate (RR\%). Best results are shown in bold; second-best results are underlined.}
\label{tab:dino_label_noise}
\begin{tabular}{lccccccc | c}
\toprule
& \multicolumn{1}{c}{MNIST} &  & \multicolumn{2}{c}{CIFAR10} &  & \multicolumn{2}{c}{CIFAR100} & \multicolumn{1}{|c}{\multirow{2}{*}{\shortstack{\textbf{Avg.}\\\textbf{RR}}}} \\
\cmidrule(lr){2-2} \cmidrule(lr){4-5} \cmidrule(lr){7-8} 
Model  & SN &  & SN & AN &  & SN & AN & \multicolumn{1}{|c}{}\\
\midrule
Mix    & 78.5 &  & 81.3 & 83.6 &  & 74.0 & 60.3 & 0.0 \\
Oracle & 99.6 &  & 98.8 & 98.7 &  & 91.0 & 91.0 & 100 \\
\cmidrule(lr){1-9}
$\text{Mix}-\tau_{r}$ & 78.1 &  & 80.4 & 83.2 &  & 71.2 & 60.0 & -5.43 \\
CF      & \underline{95.8} &  & \underline{93.0} & \underline{94.5} &  & \underline{82.4} & \underline{72.4} & \underline{61.94} \\
CUTS    & \textbf{98.8} &  & \textbf{95.4} & \textbf{94.6} &  & \textbf{84.0} & \textbf{78.6} & \textbf{73.52} \\
\bottomrule
\end{tabular}
\end{table}

\subsection{Poison Trigger}
\label{sup:subsec:extended_results_poison}
The detailed results of applying the correction methods under poison trigger corruption are reported for randomly initialized ResNet18 in \cref{tab:poison_resnet18}, for ViT initialized with \texttt{ImageNet1Kv1} weights in \cref{tab:poison_vit}, and for classifiers with a DINOv3 backbone in \cref{tab:poison_dino}. As shown, CUTS outperforms Potion across all architecture–initialization settings. We observed that the Potion algorithm is unstable in the source-free setting and often fails to remove the trigger from the corrupted model using the proxy dataset, even when provided a large trial budget (500 iterations). In some runs, Potion completely fails and severely degrades model utility, resulting in a low average PUS for this baseline, whereas CUTS remains robust across architectures and corruption rates $\eta$.

\begin{table*}[ht]
\centering
\caption{Results for randomly initialized ResNet18 models trained under poison trigger corruption. Columns show utility (UT\%) and attack success rate (ASR\%). The rightmost column shows the average poison unlearning score (PUS). Best results (PUS) are in bold; second-best are underlined.}
\label{tab:poison_resnet18}
\scriptsize
\renewcommand{\arraystretch}{1.2}
\setlength{\tabcolsep}{3.5pt}
\begin{tabular}{lccccccc cccccc c cccccc|c}
\toprule
& \multicolumn{6}{c}{MNIST} & & \multicolumn{6}{c}{CIFAR10} & & \multicolumn{6}{c}{CIFAR100} & \multicolumn{1}{|c}{\multirow{2}{*}{\shortstack{\textbf{Avg.}\\\textbf{PUS}}}} \\
\cmidrule(lr){2-7} \cmidrule(lr){9-14} \cmidrule(lr){16-21}
& \multicolumn{2}{c}{2\%} & \multicolumn{2}{c}{10\%} & \multicolumn{2}{c}{20\%} & & \multicolumn{2}{c}{2\%} & \multicolumn{2}{c}{10\%} & \multicolumn{2}{c}{20\%} & & \multicolumn{2}{c}{2\%} & \multicolumn{2}{c}{10\%} & \multicolumn{2}{c}{20\%}  & \multicolumn{1}{|c}{\textbf{}}\\
\cmidrule(lr){2-3} \cmidrule(lr){4-5} \cmidrule(lr){6-7} \cmidrule(lr){9-10} \cmidrule(lr){11-12} \cmidrule(lr){13-14} \cmidrule(lr){16-17} \cmidrule(lr){18-19} \cmidrule(lr){20-21} 
Model & UT & ASR & UT & ASR & UT & ASR & & UT & ASR & UT & ASR & UT & ASR & & UT & ASR & UT & ASR & UT & ASR & \\
\midrule
Mix & 99.7 & 100 & 99.6 & 100 & 99.7 & 100 & & 90.3 & 100 & 89.9 & 100 & 89.2 & 100 & & 69.7 & 100 & 67.5 & 100 & 66.6 & 100 & 0.0 \\
Oracle & 99.7 & 0.0 & 99.7 & 0.0 & 99.7 & 0.0 & & 90.2 & 0.0 & 89.8 & 0.0 & 89.0 & 0.0 & & 68.8 & 0.0 & 68.1 & 0.0 & 66.9 & 0.0 & 85.76 \\
\cmidrule(lr){1-22}
$\text{Mix}{-}\tau_{r}$ & 16.4 & 89.2 & 11.3 & 0.2 & 12.2 & 8.8 & & 53.3 & 94.0 & 60.4 & 77.7 & 72.1 & 49.4 & & 45.8 & 91.0 & 37.7 & 82.2 & 50.6 & 90.7 & 10.33\\
CF & 99.6 & 100 & 99.6 & 100 & 99.5 & 100 & & 89.7 & 100 & 89.1 & 100 & 87.4 & 100 & & 68.2 & 100 & 67.2 & 100 & 65.7 & 100 & 0.0 \\
Potion & \textbf{99.1} & \textbf{1.6} & \underline{46.8} & \underline{0.7} & \underline{11.4} & \underline{0.1} & & \textbf{85.9} & \textbf{0.9} & \textbf{84.8} & \textbf{1.3} & \textbf{77.5} & \textbf{1.2} & & \textbf{67.5} & \textbf{1.1} & \textbf{65.6} & \textbf{0.0} & \textbf{64.4} & \textbf{0.0} & \underline{66.38}\\
CUTS & \underline{86.1} & \underline{1.3} & \textbf{74.0} & \textbf{1.7} & \textbf{95.6} & \textbf{2.0} & & \underline{69.2} & \underline{0.9} & \underline{62.2} & \underline{1.1} & \underline{73.1} & \underline{0.9} & & \underline{58.8} & \underline{1.0} & \underline{52.8} & \underline{0.6} & \underline{60.1} & \underline{1.3} & \textbf{69.32}\\
\bottomrule
\end{tabular}
\end{table*}

\begin{table}[ht]
\centering
\caption{Results for ViT-B/16 model initialized with \texttt{ImageNet1Kv1} weights trained under poison trigger corruption. Columns show utility (UT\%) and attack success rate (ASR\%). The rightmost column shows the average poison unlearning score (PUS). Best results (PUS) are in bold; second-best are underlined.}
\label{tab:poison_vit}
\scriptsize
\renewcommand{\arraystretch}{1.2}
\setlength{\tabcolsep}{1.6pt}
\begin{tabular}{lcccccccccccc|c}
\toprule
& \multicolumn{6}{c}{CIFAR10} & \multicolumn{6}{c}{CIFAR100} & \multicolumn{1}{|c}{\multirow{2}{*}{\shortstack{\textbf{Avg.}\\\textbf{PUS}}}} \\
\cmidrule(lr){2-7} \cmidrule(lr){8-13}
& \multicolumn{2}{c}{2\%} & \multicolumn{2}{c}{10\%} & \multicolumn{2}{c}{20\%} & \multicolumn{2}{c}{2\%} & \multicolumn{2}{c}{10\%} & \multicolumn{2}{c}{20\%} & \multicolumn{1}{|c}{\textbf{}}\\
\cmidrule(lr){2-3} \cmidrule(lr){4-5} \cmidrule(lr){6-7} \cmidrule(lr){8-9} \cmidrule(lr){10-11} \cmidrule(lr){12-13}
Model & UT & ASR & UT & ASR & UT & ASR & UT & ASR & UT & ASR & UT & ASR &  \\
\midrule
Mix & 95.8 & 100 & 95.6 & 100 & 95.2 & 100 & 82.5 & 100 & 81.4 & 100 & 80.4 & 100 & 0.0 \\
Oracle & 95.9 & 0.0 & 95.8 & 0.0 & 95.7 & 0.0 & 82.7 & 0.0 & 82.6 & 0.0 & 81.8 & 0.0 & 89.08\\
\cmidrule(lr){1-14}
$\text{Mix}{-}\tau_{r}$ & 95.9 & 100 & 95.6 & 100 & 95.2 & 100 & 82.4 & 100 & 81.4 & 100 & 80.4 & 100 & 0.0\\
CF & 95.5 & 100 & 95.5 & 100 & 94.9 & 100 & 82.0 & 100 & 80.7 & 100 & 80.4 & 100 & 0.0 \\
Potion & \textbf{95.3} & \textbf{2.3} & \underline{18.2} & \underline{1.0} & \underline{12.3} & \underline{83.3} & \textbf{82.0} & \textbf{0.8} & \textbf{79.4} & \textbf{2.4} & \textbf{73.0} & \textbf{0.6} & \underline{57.41}\\
CUTS & \underline{90.8} & \underline{2.1} & \textbf{77.7} & \textbf{2.2} & \textbf{88.2} & \textbf{2.0} & \underline{79.5} & \underline{2.3} & \underline{73.0} & \underline{1.3} & \underline{68.3} & \underline{1.7} & \textbf{78.04}\\
\bottomrule
\end{tabular}
\end{table}

\begin{table}[ht]
\centering
\caption{Results for DINOv3 model trained under poison trigger corruption. Columns show utility (UT\%) and attack success rate (ASR\%). The rightmost column shows the average poison unlearning score (PUS). Best results (PUS) are in bold; second-best are underlined.}
\label{tab:poison_dino}
\scriptsize
\renewcommand{\arraystretch}{1.2}
\setlength{\tabcolsep}{1.6pt}
\begin{tabular}{lcccccccccccc|c}
\toprule
& \multicolumn{6}{c}{CIFAR10} & \multicolumn{6}{c}{CIFAR100} & \multicolumn{1}{|c}{\multirow{2}{*}{\shortstack{\textbf{Avg.}\\\textbf{PUS}}}} \\
\cmidrule(lr){2-7} \cmidrule(lr){8-13}
& \multicolumn{2}{c}{2\%} & \multicolumn{2}{c}{10\%} & \multicolumn{2}{c}{20\%} & \multicolumn{2}{c}{2\%} & \multicolumn{2}{c}{10\%} & \multicolumn{2}{c}{20\%} & \multicolumn{1}{|c}{\textbf{}}\\
\cmidrule(lr){2-3} \cmidrule(lr){4-5} \cmidrule(lr){6-7} \cmidrule(lr){8-9} \cmidrule(lr){10-11} \cmidrule(lr){12-13}
Model & UT & ASR & UT & ASR & UT & ASR & UT & ASR & UT & ASR & UT & ASR &  \\
\midrule
Mix & 98.6 & 100 & 98.7 & 100 & 98.5 & 100 & 89.3 & 100 & 89.4 & 100 & 89.0 & 100 & 0.0\\
Oracle & 98.5 & 0.0 & 98.6 & 0.0 & 98.4 & 0.0 & 89.3 & 0.0 & 89.2 & 0.0 & 88.7 & 0.0 & 93.80 \\
\cmidrule(lr){1-14}
$\text{Mix}{-}\tau_{r}$ & 98.6 & 100 & 98.6 & 100 & 98.5 & 100 & 89.2 & 99.6 & 89.2 & 100 & 88.9 & 100 & 0.06 \\
CF & 98.4 & 100 & 98.5 & 100 & 98.4 & 100 & 88.6 & 99.6 & 88.9 & 99.9 & 88.2 & 100 & 0.07\\
Potion & \textbf{98.4} & \textbf{0.8} & \underline{27.1} & \underline{0.9} & \underline{10.1} & \underline{0.0} & \textbf{87.1} & \textbf{0.0} & \underline{80.3} & \underline{0.3} & \underline{33.4} & \underline{0.4} & \underline{55.83}\\
CUTS & \underline{98.1} & \underline{0.6} & \textbf{98.4} & \textbf{1.9} & \textbf{98.0} & \textbf{1.8} & \underline{85.8} & \underline{0.5} & \textbf{85.1} & \textbf{1.4} & \textbf{72.8} & \textbf{1.5} & \textbf{88.55}\\
\bottomrule
\end{tabular}
\end{table}

\section{Extended Ablation and Analysis}
\label{sup:sec:extended_analysis}
\subsection{Size of the Proxy Dataset}
\label{sup:subsec:extended_analysis_proxy_size}
In order to assess the effect of the proxy dataset size ($|\mathcal{D}_\text{proxy}|$), we repeated the CLIP experiments on MNIST, CIFAR10, and CIFAR100 under symmetric label noise at $\eta\!\in\!\{40\%,60\%\}$, holding out $10\%$ of the original training split as $\mathcal{D}_\text{proxy}$ (instead of the $2\%$ default used in \cref{tab:clip_sym_asym}). Empirically, enlarging $\mathcal{D}_\text{proxy}$ produces a larger weight displacement during proxy fine-tuning and therefore a proxy task vector with a higher $\ell_2$ norm $\|\tau_p\|$ (see \cref{tab:clip_sym_proxy_size}). Because the estimated corruption vector is $\hat{\tau}_c = \hat{\alpha}^\ast \,\tau_p$,
the estimated scaling coefficient $\hat{\alpha}^\ast$ decreases as $\|\tau_p\|$ grows (the dependence is monotone but not strictly proportional). For a more controlled comparison of recovery and coefficients in \cref{tab:clip_sym_proxy_size}, the ideal protocol would be to fix $\mathcal{D}_\text{train}$ and draw both proxy subsets (2\% and 10\%) from the same 10\% holdout. Due to compute constraints, our 2\% setting reuses earlier runs where the remaining 98\% formed $\mathcal{D}_\text{train}$. This discrepancy should have negligible effects on $\|\tau_p\|$ (which is dominated by proxy size) but can slightly shift RR and the selected $\hat{\alpha}^\ast$.

\begin{table*}[ht]
\centering
\caption{Comparison between the effect of $|\mathcal{D}_\text{proxy}|$ on the norm of $\tau_p$, value of $\hat{\alpha}^\ast$, and the recovery rate (RR\%) of CUTS, for CLIP models trained on MNIST, CIFAR10, and CIFAR100 under symmetric label noise at $\eta \in \{40, 60\}$.}
\label{tab:clip_sym_proxy_size}
\scriptsize
\renewcommand{\arraystretch}{1.2}
\centering
\setlength{\tabcolsep}{3.5pt}
\begin{tabular}{lccccccccccccccccccccc}
\toprule
& \multicolumn{6}{c}{MNIST} & & \multicolumn{6}{c}{CIFAR-10} & & \multicolumn{6}{c}{CIFAR-100} \\
\cmidrule(lr){2-7} \cmidrule(lr){9-14} \cmidrule(lr){16-21} 
& \multicolumn{3}{c}{40\%} & \multicolumn{3}{c}{60\%} & & \multicolumn{3}{c}{40\%} & \multicolumn{3}{c}{60\%} & & \multicolumn{3}{c}{40\%} & \multicolumn{3}{c}{60\%}\\
\cmidrule(lr){2-4} \cmidrule(lr){5-7} \cmidrule(lr){9-11} \cmidrule(lr){12-14} \cmidrule(lr){16-18} \cmidrule(lr){19-21} 
$|\mathcal{D}_\text{proxy}|$ 
  & $\left\|\tau_p \right\|$ & $\hat{\alpha}^\ast$ & RR 
  & $\left\|\tau_p \right\|$ & $\hat{\alpha}^\ast$ & RR 
  & 
  & $\left\|\tau_p \right\|$ & $\hat{\alpha}^\ast$ & RR 
  & $\left\|\tau_p \right\|$ & $\hat{\alpha}^\ast$ & RR 
  &
  & $\left\|\tau_p \right\|$ & $\hat{\alpha}^\ast$ & RR 
  & $\left\|\tau_p \right\|$ & $\hat{\alpha}^\ast$ & RR \\
\midrule
$2\%$ 
  & 0.42 & 1.5  & 90.8 & 0.40 & 3.0  & 93.4 
  & 
  & 0.31 & 1.4  & 86.5 & 0.29 & 2.0  & 85.8 
  & 
  & 0.41 & 1.35 & 61.1 & 0.38 & 2.0  & 71.2\\
\cmidrule(lr){1-21}
$10\%$ 
  & 1.11 & 1.4  & 90   & 1.06 & 1.9  & 91 
  & 
  & 0.76 & 1.0  & 84.9 & 0.91 & 1.3  & 83.6 
  & 
  & 0.98 & 0.85 & 63.32 & 0.94 & 1.35 & 70.1\\  
\bottomrule
\end{tabular}
\end{table*}


\subsection{Cosine Similarity of the Task Vectors}
\label{sup:subsec:extended_analysis_cosine_sims}

In this section we investigate whether task vector directions carry meaningful information. We compute cosine similarities among the proxy task vector $\tau_p$, the random task vector $\tau_r$, the joint-training vector $\tau_{\text{mix}}$, and the vector associated with the catastrophic forgetting baseline,
$\tau_{\text{CF}} = \theta_{\text{CF}} {-} \theta_{\text{mix}}$.
For symmetric label noise we also report two proxy vectors built from the same proxy dataset input images but with different random labels (obtained using different noise seeds), $\tau_p^1$ and $\tau_p^2$, and their average $\tau_p^{\text{avg}}$ (see \cref{sup:subsec:training_details_CUTS}). To compute similarities, we flatten all parameters in each task vector and compute the cosine between the resulting vectors.

\begin{figure}[ht]
    \centering
    \includegraphics[width=\linewidth]{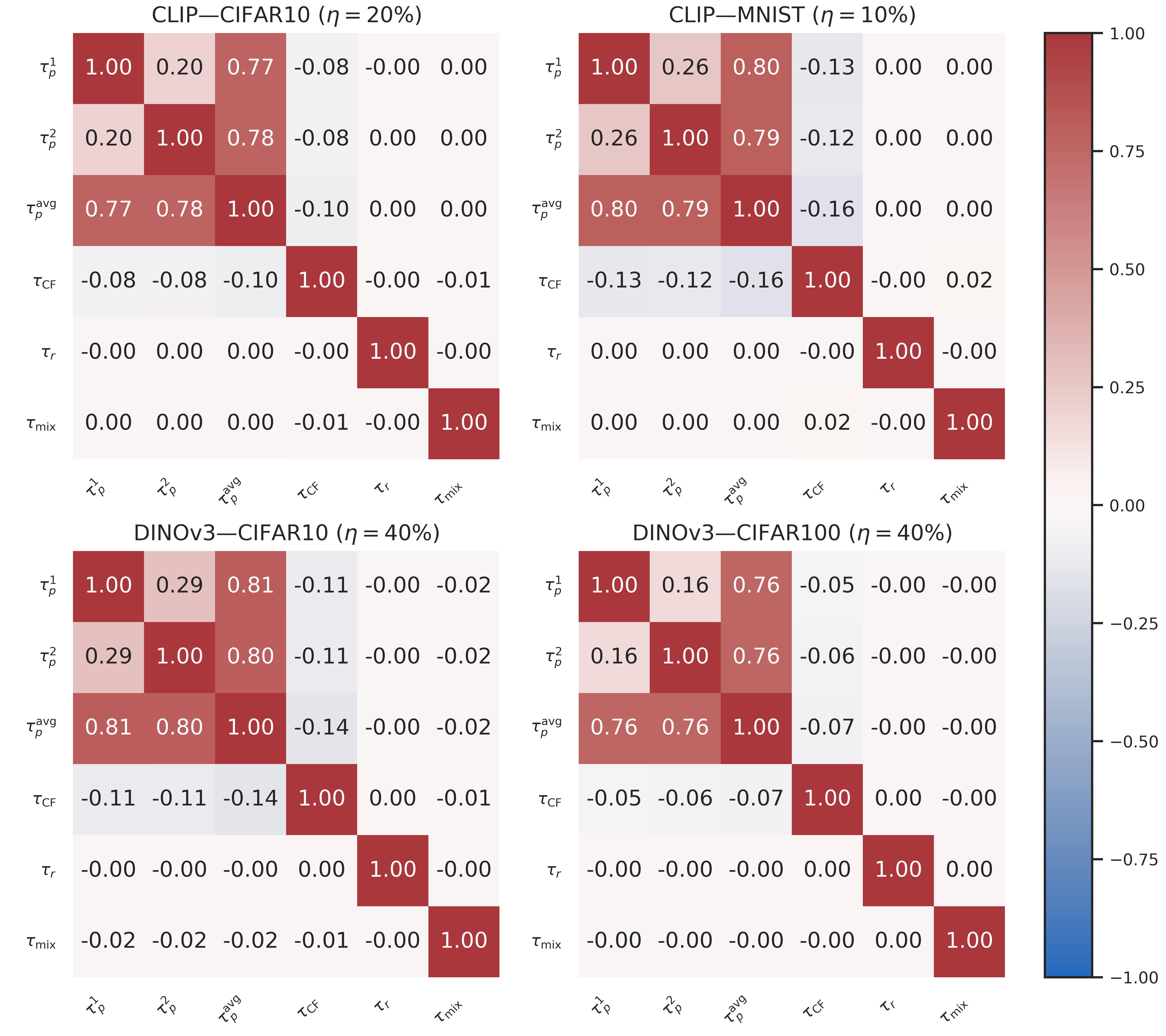}
    \caption{Cosine similarity between task vectors for experiments on symmetric label noise.}
    \label{fig:task_sim_grid_sym}
\end{figure}

The similarity matrices for four experiments with symmetric label noise are shown in \cref{fig:task_sim_grid_sym}. Across all settings, $\tau_p^1$ and $\tau_p^2$ have positive cosine similarity, and each is strongly aligned with their average $\tau_p^{\text{avg}}$. This indicates that different noise realizations lead to very similar proxy directions in weight space. In particular, the proxy vectors associated with symmetric label noise appear to concentrate around a shared direction that is stable across label noise seeds, consistent with our interpretation of $\tau_p$ as capturing the (symmetric) label corruption task direction. Moreover, the cosine similarity between $\tau_p^{\text{avg}}$ and $\tau_{\text{CF}}$ is consistently negative (around $-0.1$), indicating a mild but systematic anti-alignment between proxy fine-tuning and catastrophic forgetting fine-tuning.

\begin{figure}[ht]
    \centering
    \includegraphics[width=\linewidth]{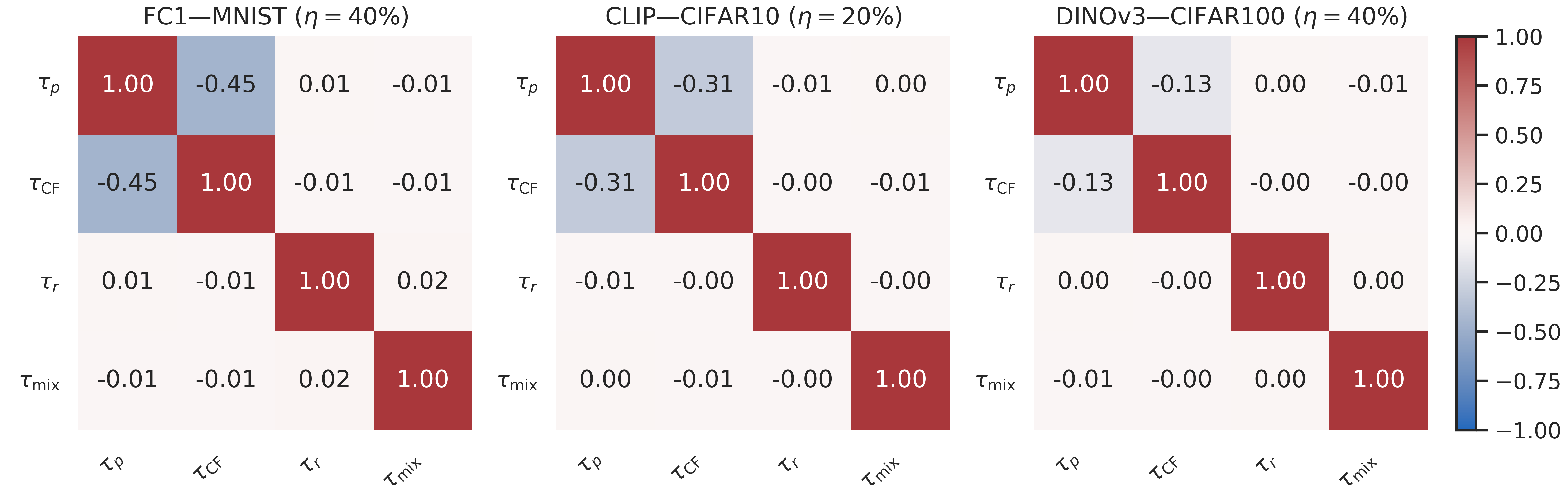}
    \caption{Cosine similarity between task vectors for experiments on asymmetric label noise.}
    \label{fig:task_sim_grid_asym}
\end{figure}

\Cref{fig:task_sim_grid_asym} reports similarity matrices for asymmetric label noise. In this case, the cosine between $\tau_p$ and $\tau_{\text{CF}}$ is more strongly negative (down to about $-0.45$), showing that the proxy direction is more clearly aligned with the negative of the catastrophic forgetting direction than in the symmetric noise case. Together with the quantitative results in \cref{tab:full_resnet_fc_noise_results}, this supports the view that CUTS provides a correction update that is closely related to the clean-label fine-tuning used by the CF baseline. Hence, when only corrupted samples are available (for instance, generated from model mistakes or domain knowledge rules), the proxy based update can act as a viable alternative to CF.

\begin{figure}[ht]
    \centering
    \includegraphics[width=\linewidth]{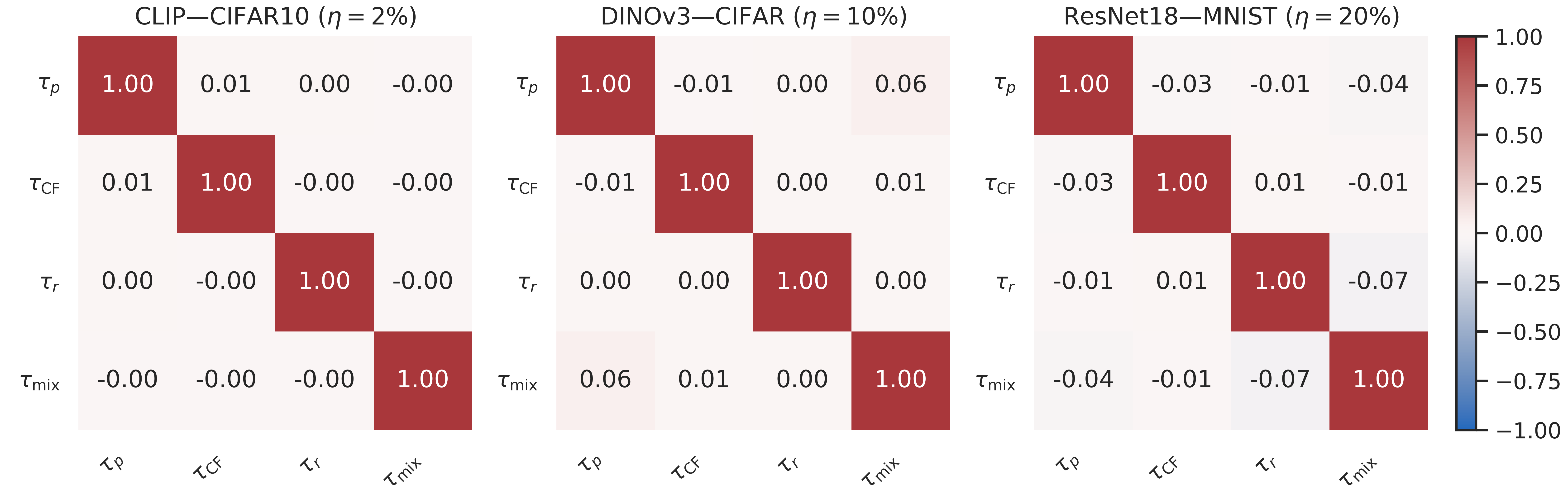}
    \caption{Cosine similarity between task vectors for experiments on poison trigger.}
    \label{fig:task_sim_grid_pois}
\end{figure}

For poison trigger corruption (\cref{fig:task_sim_grid_pois}), the task vectors $\tau_{\text{CF}}$ and $\tau_p$ are nearly orthogonal in all experiments. This shows that the update induced by proxy fine-tuning points in a direction that is essentially independent of the direction taken by the CF baseline when fine-tuning on clean data. Combined with the empirical observation that the CF baseline fails to reliably remove poison triggers (see \cref{tab:poison_resnet18}), this suggests that, in our setting, gradients from clean samples alone are not sufficient to discover the trigger removal direction identified by the proxy task.

Another observation is that, for models initialized from strong pre-trained backbones (CLIP and DINOv3), the random direction $\tau_r$ is almost orthogonal to all other task vectors. This suggests that, in the high-dimensional weight space of these models, only a small set of structured directions substantially alter the model behavior, while perturbations along a random direction tend to leave performance largely unchanged, consistent with the behavior of the Mix${-}\tau_r$ baseline reported in our tables. In contrast, for randomly initialized classifiers, moving along random directions has a much stronger effect on performance (cf.\ \cref{tab:full_resnet_fc_noise_results,tab:poison_resnet18}), indicating that the weight space of such models is less geometrically disentangled.

Finally, we observe that both $\tau_p$ and $\tau_{\text{CF}}$ are nearly orthogonal to $\tau_{\text{mix}}$ in all corruption types. Therefore, both the CUTS correction direction defined by $-\tau_p$ and the CF fine-tuning direction defined by $\tau_{\text{CF}}$ primarily move the parameters in directions that lie outside the joint-training direction defined by $\tau_{\text{mix}}$.

\subsection{Corrected ``Clean'' Samples for Uncorrupted Training Datasets}
\label{sup:subsec:extended_analysis_clean_forgotten}
\begin{figure}[ht]
    \centering
     
    \includegraphics[width=\linewidth]{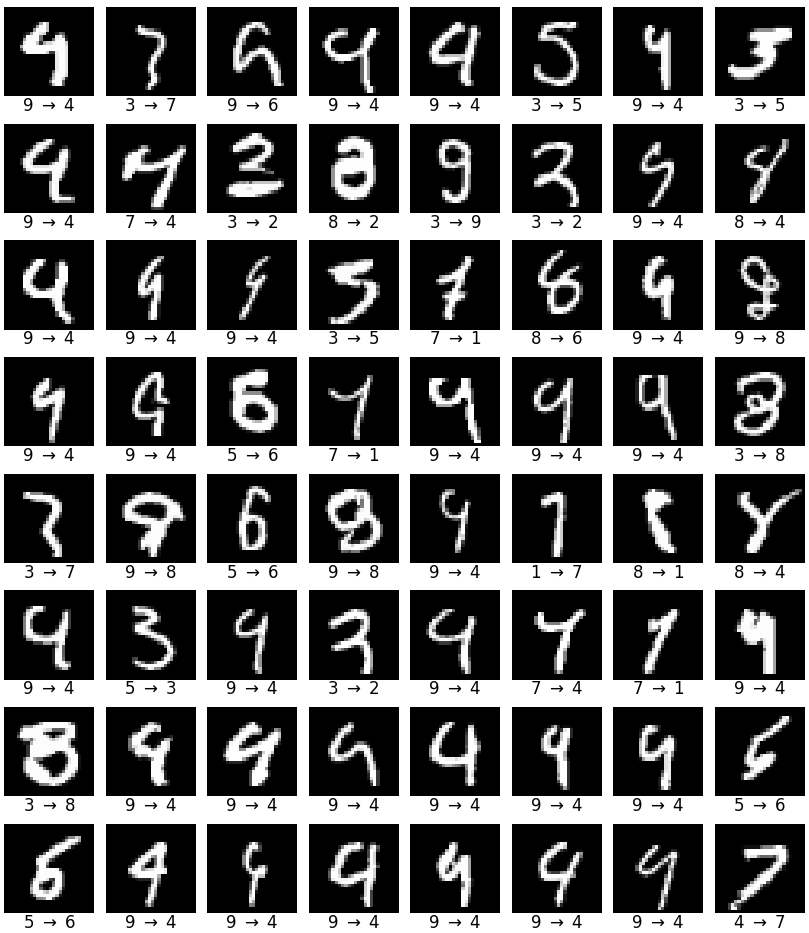}
    \caption{For a CLIP model trained on the uncorrupted MNIST training split, the examples above show the first $64$ samples whose predictions flip away from their ground-truth labels when increasing $\alpha$ along $-\tau_p$. The proxy task vector $\tau_p$ is built using the MNIST test split corrupted with symmetric label noise. Captions show \emph{ground-truth $\rightarrow$ predicted} after the flip.} 
    \label{fig:forgotten_samples_clean_mnist}
\end{figure}

\begin{figure}[ht]

    \centering
     
    \includegraphics[width=1.\linewidth]{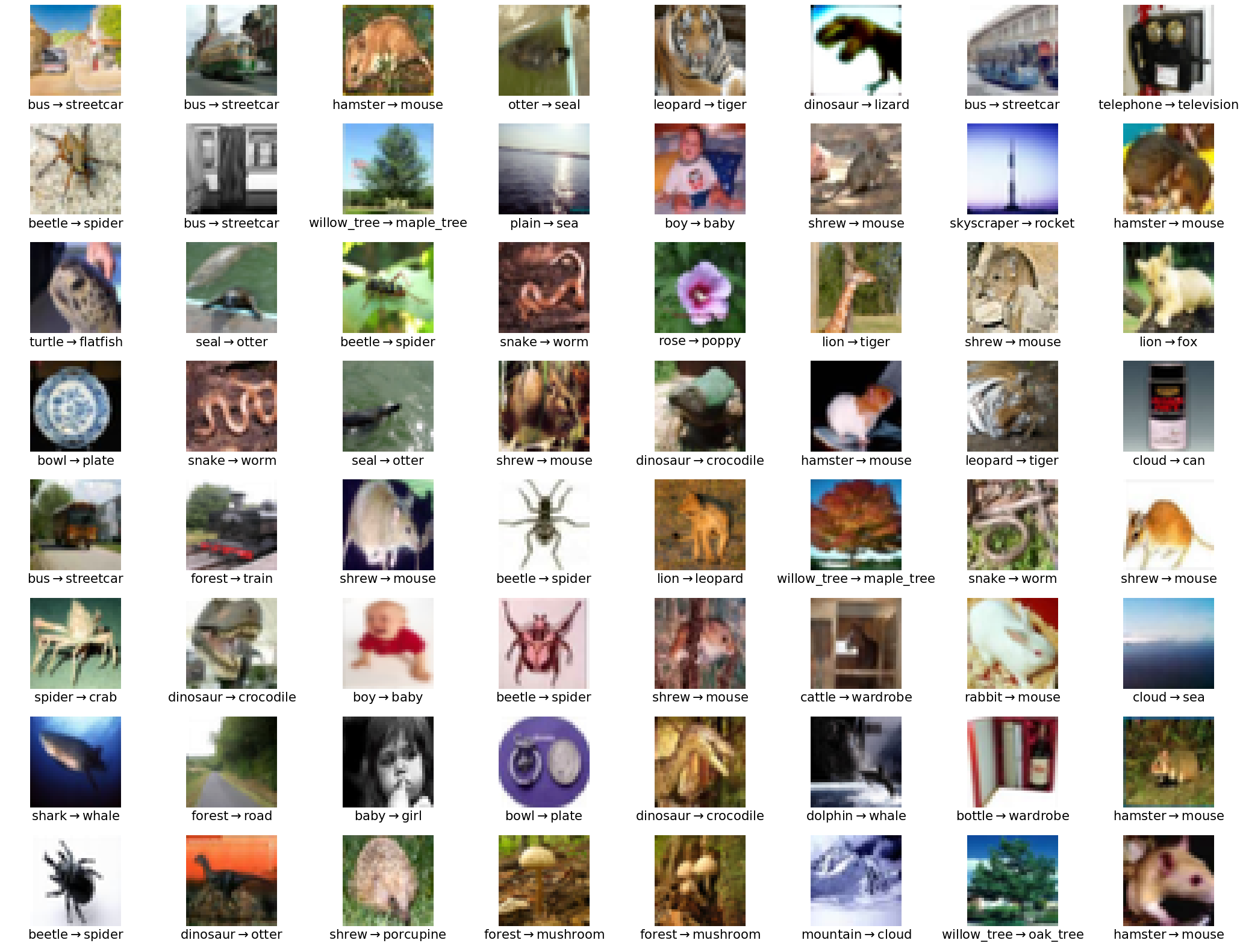}
    \caption{For a CLIP model trained on the uncorrupted CIFAR100 training split, the examples above show the first $64$ samples whose predictions flip away from their ground-truth labels when increasing $\alpha$ along $-\tau_p$. The proxy task vector $\tau_p$ is built using the CIFAR100 test split corrupted with symmetric label noise. Captions show \emph{ground-truth $\rightarrow$ predicted} after the flip.} 
    \label{fig:forgotten_samples_clean_cifar100}
\end{figure}

We repeated the analysis regarding first flips during correction on models trained on the original (uncorrupted) MNIST and CIFAR100 training sets, without any artificially injected label noise. To construct the corruption task vector, we used samples from the corresponding test sets as proxy datasets. We then recorded the first $64$ training examples whose predictions flipped away from their training labels. The resulting sets are shown in \cref{fig:forgotten_samples_clean_mnist} (MNIST) and \cref{fig:forgotten_samples_clean_cifar100} (CIFAR100). Qualitatively, many of the samples are likely mislabeled or ambiguous, and the rest are atypical or low-support instances. These results show that even widely used benchmark datasets contain mislabeled or highly ambiguous samples.

\begin{table}[p]
\centering
\caption{Details of the training hyperparameters for different experiments and models. For CLIP and DINOv3, the \textit{Total Iterations} values denote the number of gradient steps, whereas for FC1, ResNet18/34/50/101, and ViT-B/16 they denote the number of epochs. For \emph{Step} schedule, we only specify the number of milestones, and the value of gamma.}
\label{tab:training_hyperparams}
\scriptsize
\renewcommand{\arraystretch}{1.2}
\setlength{\tabcolsep}{2pt}
\begin{tabular}{lcccccccccc}
\toprule
\multicolumn{3}{l}{Experiment} & Model & {\makecell{Total\\Iterations}} & {\makecell{Max\\LR}} & {\makecell{LR\\Schedule}} & {\makecell{Warmup/\\Milestones}} &  {\makecell{Hold/\\Gamma}} & {\makecell{Batch\\Size}} \\
\midrule
\multirow{12}{*}{\rotatebox{90}{\large CLIP}} & \multirow{8}{*}{\rotatebox{90}{\textbf{SN/AN}}}  & \multirow{4}{*}{MNIST} & Mix & 13000 & 1e-5 & Cosine & 300  & 10000  & 256 \\
& & & Oracle& 3000 & 1e-5 & Cosine & 300  & 0  & 256 \\
& & & Proxy & 800  & 1e-5 & Cosine & 200  & 200  & 256\\
& & & CF & 500  & 1e-5 & Cosine & 100  & 200  & 256\\
\cmidrule(l){3-10}
& & \multirow{4}{*}{\makecell{CIFAR10/ \\ CIFAR100}} & Mix & 6000  & 1e-5 & Cosine & 300  & 0  & 256 \\
& & & Oracle & 3000  & 1e-5 & Cosine & 300  & 0  & 256 \\
& & & Proxy & 800  & 1e-5 & Cosine & 200  & 200  & 256\\
& & & CF & 500  & 1e-5 & Cosine & 100  & 200  & 256\\
\cmidrule(l){2-10}

& \multirow{4}{*}{\rotatebox{90}{\textbf{PT}}}&\multirow{4}{*}{\makecell{MNIST/ \\ CIFAR10/ \\ CIFAR100}} & Mix & 6000  & 1e-5 & Cosine & 300  & 2700  & 256 \\
& & & Oracle & 3000  & 1e-5 & Cosine & 300  & 700  & 256 \\
& & & Proxy & 100  & 1e-5 & Cosine & 30  & 20  & 256\\
& & & CF & 100  & 1e-5 & Cosine & 30  & 30  & 256\\

\midrule
\multirow{15}{*}{\rotatebox{90}{\large DINOv3}} & \multirow{11}{*}{\rotatebox{90}{\textbf{SN/AN}}}  & \multirow{3}{*}{\textbf{Clothing1M}} & Mix & 150000 & 1e-5 & Cosine & 1000  & 79000  & 256 \\
& & & Proxy & 500  & 1e-5 & Cosine & 100  & 300  & 256\\
& & & CF & 500  & 1e-5 & Cosine & 100  & 200  & 256\\
\cmidrule(l){3-10}
& & \multirow{4}{*}{MNIST} & Mix & 15000 & 5e-5 & Cosine & 300  & 8000  & 128 \\
& & & Oracle & 4000 & 5e-5 & Cosine & 300  & 2000  & 128 \\
& & & Proxy & 600  & 1e-5 & Cosine & 100  & 300  & 128\\
& & & CF & 500  & 1e-5 & Cosine & 100  & 200  & 128\\
\cmidrule(l){3-10}
& & \multirow{4}{*}{\makecell{CIFAR10/ \\ CIFAR100}} & Mix & 12000 & 5e-5 & Cosine & 300  & 8000  & 128 \\
& & & Oracle & 4000 & 5e-5 & Cosine & 300  & 2000  & 128 \\
& & & Proxy & 600  & 1e-5 & Cosine & 100  & 300  & 128\\
& & & CF & 500  & 1e-5 & Cosine & 100  & 200  & 128\\
\cmidrule(l){2-10}

& \multirow{4}{*}{\rotatebox{90}{\textbf{PT}}}&\multirow{4}{*}{\makecell{CIFAR10/ \\ CIFAR100}} & Mix & 6000  & 1e-5 & Cosine & 300  & 2700  & 256 \\
& & & Oracle & 4000  & 1e-5 & Cosine & 300  & 1700  & 256 \\
& & & Proxy & 100  & 1e-5 & Cosine & 30  & 20  & 256\\
& & & CF & 100  & 1e-5 & Cosine & 30  & 30  & 256\\

\midrule
\multirow{4}{*}{\rotatebox{90}{\large FC1}} & \multirow{4}{*}{\rotatebox{90}{\textbf{SN/AN}}}  & \multirow{4}{*}{MNIST} & Mix & 200 & 5e-4 & Cosine & 0  & 0  & 1024 \\
& & & Oracle & 200 & 5e-4 & Cosine & 0  & 0  & 1024 \\
& & & Proxy & 300  & 1e-4 & Step & 5  & 0.8  & 1024\\
& & & CF & 30  & 5e-5 & Step & 5  & 0.8  & 1024\\

\midrule
\multirow{12}{*}{\rotatebox{90}{\large ResNet18/34/50/101}} & \multirow{8}{*}{\rotatebox{90}{\textbf{SN/AN}}}  & \multirow{4}{*}{\makecell{CIFAR-10/ \\ CIFAR-100 \\ RND Init}} & Mix & 150 & 1e-3 & Cosine & 0  & 0  & 256 \\
& & & Oracle & 150 & 1e-3 & Cosine & 0  & 0  & 256 \\
& & & Proxy & 120  & 5e-4 & Step & 6  & 0.7  & 256\\
& & & CF & 80  & 5e-5 & Step & 4  & 0.9  & 256\\
\cmidrule(l){3-10}
& & \multirow{4}{*}{\makecell{CIFAR10/ \\ CIFAR100 \\ Pretrained}} & Mix & 120 & 5e-4 & Cosine & 0  & 0  & 256 \\
& & & Oracle & 120 & 5e-4 & Cosine & 0  & 0  & 256 \\
& & & Proxy & 120  & 5e-4 & Step & 6  & 0.7  & 256\\
& & & CF & 80  & 5e-5 & Step & 4  & 0.9  & 256\\
\cmidrule(l){2-10}

& \multirow{4}{*}{\rotatebox{90}{\textbf{PT}}}&\multirow{4}{*}{\makecell{MNIST/ \\CIFAR10/ \\ CIFAR100}} & Mix & 120  & 1e-3 & Cosine & 0  & 0  & 256 \\
& & & Oracle & 120  & 1e-3 & Cosine & 0  & 0  & 256 \\
& & & Proxy & 200  & 1e-4 & Cosine & 30  & 120  & 256\\
& & & CF & 30  & 5e-5 & Cosine & 10  & 10  & 256\\

\midrule
\multirow{8}{*}{\rotatebox{90}{\large ViT-B/16}} & \multirow{4}{*}{\rotatebox{90}{\textbf{SN/AN}}}  & \multirow{4}{*}{\makecell{CIFAR10/ \\ CIFAR100}} & Mix & 100 & 5e-4 & Cosine & 10  & 50  & 256 \\
& & & Oracle & 70 & 5e-4 & Cosine & 10  & 30  & 256 \\
& & & Proxy & 60  & 1e-4 & Cosine & 10  & 20  & 256\\
& & & CF & 60  & 1e-4 & Cosine & 10  & 20  & 256\\
\cmidrule(l){2-10}

& \multirow{4}{*}{\rotatebox{90}{\textbf{PT}}}&\multirow{4}{*}{\makecell{MNIST/ \\CIFAR10/ \\ CIFAR100}} & Mix & 80  & 5e-4 & Cosine & 10  & 30  & 256 \\
& & & Oracle & 60  & 5e-4 & Cosine & 10  & 20  & 256 \\
& & & Proxy & 60  & 1e-4 & Cosine & 10  & 20  & 256\\
& & & CF & 60  & 1e-4 & Cosine & 10  & 20  & 256\\

\bottomrule
\end{tabular}
\end{table}

\end{document}